\PassOptionsToPackage{numbers}{natbib}
\documentclass{article}
\usepackage[most]{tcolorbox}
\usepackage{adjustbox}
\usepackage{fvextra}

\usepackage[preprint]{neurips_2024}

\usepackage[utf8]{inputenc} 
\usepackage[T1]{fontenc}    
\usepackage{hyperref}       
\usepackage{url}            
\usepackage{booktabs}       
\usepackage{tabularx}
\usepackage{makecell}
\usepackage{amsfonts}       
\usepackage{nicefrac}       
\usepackage{microtype}      
\usepackage{xcolor}         
\usepackage{enumitem}
\usepackage[utf8]{inputenc}
\usepackage{setspace}
\usepackage{amssymb} 
\usepackage{pifont}  
\usepackage[most]{tcolorbox}

\usepackage{natbib}

\usepackage{amsmath}
\usepackage{graphicx}

\makeatletter
\renewcommand{\@noticestring}{}
\makeatother

\title{ASTRA: Automated Synthesis of agentic Trajectories and Reinforcement Arenas}

\begin{document}
\makeatletter
\renewcommand\@fnsymbol[1]{\ifcase#1\or 1\else\@arabic{#1}\fi}
\makeatother

\author{
  \raisebox{-0.4em}{\includegraphics[height=1.5em]{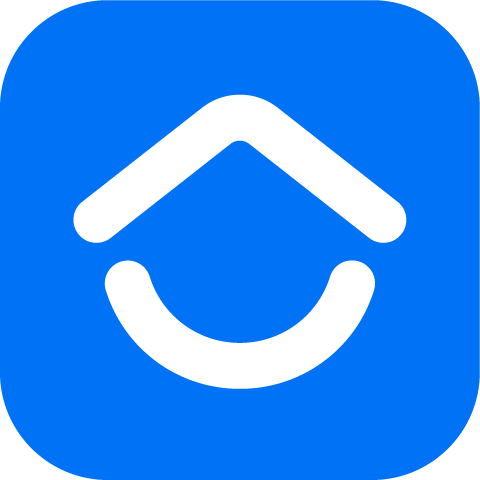}}
  \hspace{0.2em}Beike Language and Intelligence\thanks{Beike Language and Intelligence (BLI). For the complete list of authors, please refer to the \nameref{sec:contribution} section.}
}
\date{}

\maketitle

\begin{abstract}
Large language models (LLMs) are increasingly used as tool-augmented agents for multi-step decision making, yet training robust tool-using agents remains challenging. Existing methods still require manual intervention, depend on non-verifiable simulated environments, rely exclusively on either supervised fine-tuning (SFT) or reinforcement learning (RL), and struggle with stable long-horizon, multi-turn learning. To address these challenges, we introduce \textbf{ASTRA}, a fully automated end-to-end framework for training tool-augmented language model agents via scalable data synthesis and verifiable reinforcement learning. ASTRA integrates two complementary components. First, a pipeline that leverages the \textbf{static topology of tool-call graphs} synthesizes diverse, structurally grounded trajectories, instilling broad and transferable tool-use competence. Second, an environment synthesis framework that captures the \textbf{rich, compositional topology of human semantic reasoning} converts decomposed question–answer traces into independent, code-executable, and rule-verifiable environments, enabling deterministic multi-turn RL. Based on this method, we develop a unified training methodology that integrates SFT with online RL using trajectory-level rewards to balance task completion and interaction efficiency. Experiments on multiple agentic tool-use benchmarks demonstrate that ASTRA-trained models achieve state-of-the-art performance at comparable scales, approaching closed-source systems while preserving core reasoning ability. We release the full pipelines, environments, and trained models at \href{https://github.com/LianjiaTech/astra}{https://github.com/LianjiaTech/astra}.
\end{abstract}

\begin{figure}[h!]
    \centering
    \includegraphics[width=0.80\linewidth]{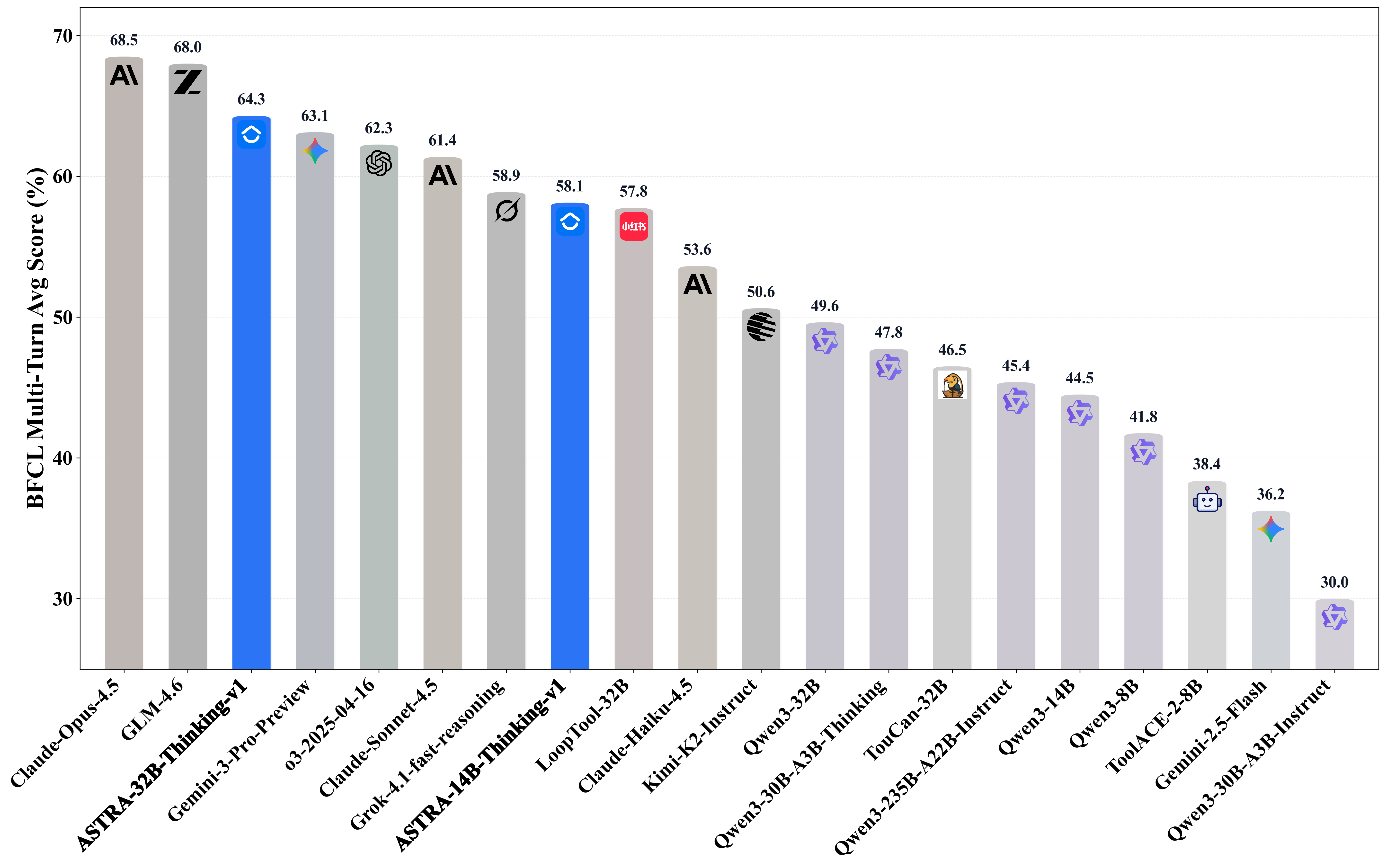}
    \caption{Comparison of Model Performance on BFCL v3 Multi-Turn.}
\end{figure}

\clearpage
\section{Introduction}

Large language models (LLMs) are increasingly deployed as \textbf{tool-augmented agents} that interact with external environments, invoke APIs, and perform multi-step decision making. By integrating reasoning with action, such agents enable applications ranging from information retrieval and data analysis to interactive dialogue systems, making tool use a core capability of modern language models.

Despite rapid progress, training robust and generalizable tool agents remains challenging. Recent work~\citep{li2025closeloopsynthesizinginfinite, zhang2025looptoolclosingdatatrainingloop} has begun to reduce human intervention by automatically synthesizing tool-use data and environments through model-driven simulation, significantly improving scalability and coverage.

However, many of these approaches rely on \textbf{LLM-simulated environments}, where tool executions, state transitions, and feedback are generated through language-model rather than explicit rules or executable backends. That is, their reinforcement learning (RL) setups are \textbf{not rule-verifiable}. This lack of verifiability fundamentally limits stable long-horizon, multi-turn online RL, where deterministic transitions and reliable reward signals are critical. Moreover, several methods~\citep{zhang2025looptoolclosingdatatrainingloop, ye2025feedbackdriventooluseimprovementslarge} generate multi-turn trajectories offline but decompose them into isolated single-step training instances, which \textbf{limits the agent’s ability to learn coherent long-horizon, multi-turn decision making}.

In addition, many existing approaches focus on only a single training regime—either supervised fine-tuning (SFT) or RL~\citep{zeng2025toolzerotrainingtoolaugmented, qian2025toolrlrewardtoollearning, liu2025toolacewinningpointsllm}. SFT-only methods lack online learning signals from environment interaction, while RL-only approaches are fundamentally constrained by the capability of the original model, limiting their effectiveness when starting from weaker initial policies.

To address these challenges, we present \textbf{ASTRA}, a fully automated, end-to-end framework for training tool-augmented language model agents via scalable data synthesis and verifiable multi-turn online reinforcement learning. ASTRA removes human intervention throughout both data construction and validation, and is fully open-sourced.

ASTRA integrates two complementary components. For SFT, we propose a trajectory synthesis pipeline that leverages the \textbf{static topology of tool-call graphs} to construct diverse multi-turn tool-use trajectories grounded in real MCP servers, and automatically scores them for quality—enabling high-quality supervised fine-tuning without manual annotation. For RL, we introduce an environment synthesis framework that captures the \textbf{rich, compositional topology of human semantic reasoning}, converting decomposed question--answer traces into independent, code-executable, and rule-verifiable environments, thereby supporting multi-turn, long-horizon RL with deterministic rewards.

Building on these components, we develop a complete training methodology for tool agents.We use SFT to learn a stronger initial policy that is better adapted to multi-turn tool interaction and then perform online, multi-turn RL over diverse synthesized environments, incorporating irrelevant-tool mixing and an F1-style trajectory-level reward to jointly optimize task completion and interaction efficiency.

\textbf{This two-stage method first broadens an agent’s tool-use competence over a static tool topology, then deepens its capability by learning within a complex semantic topology.} As a result, ASTRA effectively balances generalization across tools with robustness in realistic, high-complexity scenarios.

Our contributions are summarized as follows:
\begin{itemize}[leftmargin=1.2em]
    \item We propose a fully automated, end-to-end data construction pipeline for tool-agent training, leveraging the static topology of tool-call graphs for multi-turn trajectory synthesis and capturing the rich, compositional topology of human semantic reasoning for QA-derived, rule-verifiable environment construction.

    \item We propose a complete training methodology consisting of (i) supervised fine-tuning for a stronger initial policy that is better adapted to multi-turn tool interaction, and (ii) multi-turn online  reinforcement learning over code-executable, rule-verifiable environments across multiple domains, enabling reliable and scalable agent training.

    \item ASTRA-trained models achieve state-of-the-art performance among the same scale on multiple agentic tool-use benchmarks, approaching closed-source systems while preserving core reasoning ability, and we make the data synthesis pipelines and trained models publicly available to support reproducibility and future research.
\end{itemize}


\section{Tool-Integrated Trajectory and Verifiable Environment Synthesis}

We first present the tool-chain-based trajectory synthesis pipeline for SFT, followed by the QA-based environment synthesis framework for RL.

\subsection{Multi-turn Trajectory Synthesis with MCP Services and Tool Emulators}
\label{sec:trajectory_synthesis}
\begin{figure}[h!]
    \centering
    \includegraphics[width=1.0\linewidth]{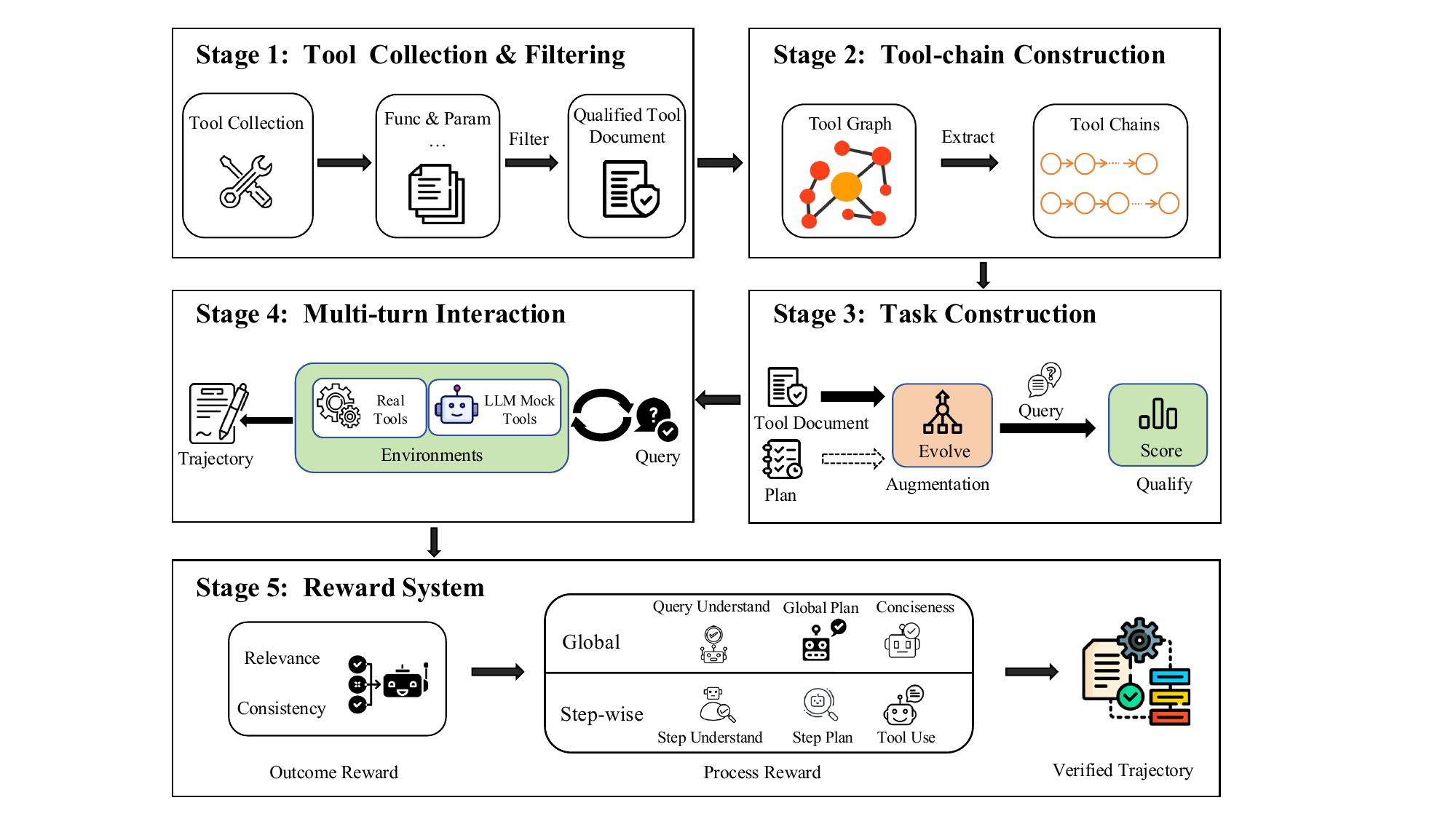}
    \caption{Overview of the Tool-Chain-Based Trajectory Synthesis Pipeline.}
    \label{fig:sft-pipeline}
\end{figure}

As illustrated in Figure~\ref{fig:sft-pipeline}, the pipeline begins with tool document collection and normalization, followed by tool-chain synthesis and validation. We then generate tasks with enhanced realism and diversity, and finally perform trajectory rollouts using an agent framework that integrates both real and simulated tools.

\subsubsection{Tool Document Collection and Filtering}
\label{subsec:tool_doc_collection}

We begin by collecting tool documents from heterogeneous sources, including (i) open MCP registries and API platforms (e.g., Smithery ~\citep{smithery2026} and RapidAPI ~\citep{rapidapi2026}), (ii) internal tool specifications, and (iii) tool documentation extracted from public datasets. Then, the tool documentations are processed in the following two stages.

\paragraph{Schema normalization.}
We convert all tools into a unified schema compatible with the OpenAI client tool-calling protocol. This normalization yields a consistent representation across tool providers, which simplifies integration during deployment and inference.

\paragraph{Grouping and filtering.}
Tool documents from different sources are first grouped by their originating service. For clarity, we refer to each group as an MCP server in the remainder of this section. We then apply quality filters to keep only groups that can support non-trivial multi-step interactions:
\begin{itemize}[leftmargin=1.2em]
  \item \textbf{Sufficient number of tools } We discard MCP servers with fewer than three tools or functions, as they rarely support meaningful multi-turn workflows.
  \item \textbf{Clear and actionable descriptions } We discard tool documents whose descriptions are too vague to determine intended functionality (e.g., missing descriptions or irreducible ambiguity after cleaning).
  \item \textbf{Convertible to the unified format } We exclude tools whose schemas cannot be reliably mapped into the OpenAI-style tool-calling format used in our normalization step.
\end{itemize}

After grouping and filtering, we retained \textbf{1,585} MCP servers, comprising \textbf{19,036} tool documents spanning \textbf{41} domains.

\subsubsection{Tool-chain Construction}
\label{subsec:tool_chain}

Formally, for an MCP server $s$, let $\mathcal{F}^{(s)}=\{f^{(s)}_1,\dots,f^{(s)}_{m_s}\}$ denote the set of tools exposed by $s$.
Each tool $f\in\mathcal{F}^{(s)}$ provides an input schema $\mathcal{I}(f)$, along with natural-language documentation (e.g., a tool description and per-argument descriptions).
In this work, we restrict composition to tools within the same MCP server and do not compose tools across servers.

\paragraph{Tool-chains as task-conditioned execution sequences.}
For each server $s$, we use an LLM to jointly synthesize (i) a possible tool-relative task and (ii) a plausible tool-chain that could be used to solve it.
A tool-chain is a length-$n$ sequence $\mathbf{c}=(f_1,\dots,f_n)$ with $f_i\in\mathcal{F}^{(s)}$.
The synthesis conditions on each tool's input schema and natural-language documentation, encouraging chains whose successive calls are supported by the task specification and information implied by earlier tools.

\paragraph{Candidate chain construction via transition-graph walks.}
For each server $s$, we run the joint synthesis procedure to obtain a multiset of tool-chains
$\mathcal{C}^{(s)}=\{\mathbf{c}_1,\dots,\mathbf{c}_N\}$, where each $\mathbf{c}_\ell=(f^{(\ell)}_1,\dots,f^{(\ell)}_{n_\ell})$ and $f^{(\ell)}_i\in\mathcal{F}^{(s)}$.

We then aggregate $\mathcal{C}^{(s)}$ into a directed transition graph $\widehat{G}^{(s)}=(V^{(s)},\widehat{E}^{(s)},w)$,
where $V^{(s)}$ contains one node per tool in $\mathcal{F}^{(s)}$, and an edge $(f_i\!\rightarrow\! f_j)\in\widehat{E}^{(s)}$ is added if the ordered pair appears consecutively in any synthesized chain.

Finally, we sample candidate tool-chains by performing length-bounded random walks on $\widehat{G}^{(s)}$ (biased by $w$),
and keep walks that satisfy basic validity constraints (e.g., maximum length and optional acyclicity).
The resulting walks constitute our final candidate chains for server $s$.

\paragraph{Dependency verification.}
We apply two checks to each sampled chain.
First, we verify inter-tool dependencies: required arguments for each tool can be supported by the task specification and fields implied by earlier tools, yielding a well-formed dependency structure.
Second, we validate task--chain coherence by filtering out chains paired with ill-posed or nonsensical tasks.
Chains failing either check are discarded.

\subsubsection{Task Construction, Augmentation, and Scoring}
\label{subsec:task_generation}

For each MCP server $s$ with tool set $\mathcal{F}^{(s)}$, we synthesize user tasks that are (i) plausible as genuine requests and (ii) solvable via tool usage provided by the server. Our pipeline combines complementary construction modes to balance realism and coverage, then applies controlled augmentation and quality-based filtering.

\paragraph{Task construction.}
We generate an initial task set $\mathcal{T}^{(s)}$ by combining two complementary sources,
$\mathcal{T}^{(s)} = \mathcal{T}^{(s)}_{\text{chain}} \cup \mathcal{T}^{(s)}_{\text{server}}$,
where the two components emphasize executability and coverage respectively:
\begin{itemize}[leftmargin=1.2em]
    \item \textbf{Chain-conditioned construction ($\mathcal{T}^{(s)}_{\text{chain}}$) }
    Given a server specification and a validated tool-chain, we condition the LLM
    \footnote{We use GLM-4.6-FP8~\citep{5team2025glm45agenticreasoningcoding} as the default LLM unless otherwise specified }
    to generate tasks whose solutions naturally follow a coherent multi-step workflow consistent with validated tool-chain.
    This setting biases generation toward tasks that correspond to executable tool interactions.
    
    \item \textbf{Server-only construction ($\mathcal{T}^{(s)}_{\text{server}}$) }
    Given only the server specification, we generate task candidates that can be solved using tools from $\mathcal{F}^{(s)}$.
    This setting promotes broader topical and linguistic coverage, reducing redundancy and overly constrained scenarios.

\end{itemize}

\paragraph{Task augmentation with consistency constraints.}
Starting from $\mathcal{T}^{(s)}$, we expand the distribution by applying an augmentation operator $\mathcal{A}(\cdot)$, yielding an augmented set
$\widetilde{\mathcal{T}}^{(s)} = \mathcal{T}^{(s)} \cup \mathcal{A}(\mathcal{T}^{(s)})$.
We instantiate $\mathcal{A}$ along three complementary axes:
\begin{itemize}[leftmargin=1.2em]
  \item \textbf{Diversity augmentation } Paraphrastic and content-varied rewrites (e.g., alternative wording, preference expressions, or contextual backgrounds) that preserve the same intent.
  \item \textbf{Complexity augmentation } Introduce additional requirements (e.g., multi-constraint preferences, implicit references, or follow-up needs) while keeping the core goal unchanged.
  \item \textbf{Persona-conditioned augmentation } Rewrite tasks under user personas (e.g., \textit{novice} vs. \textit{expert}, \textit{concise} vs. \textit{verbose}) to cover diverse communication patterns.
\end{itemize}

To mitigate distribution drift, for each original task $t\in\mathcal{T}^{(s)}$ and its augmented variant $\tilde{t}\in\mathcal{A}(\mathcal{T}^{(s)})$, we enforce \textbf{language consistency} by requiring $\mathrm{lang}(\tilde{t})=\mathrm{lang}(t)$, where $\mathrm{lang}(\cdot)$ denotes the task's language category.
Furthermore, we constrain augmentation to preserve the semantic intent and logical requirements of the original task.

\paragraph{Task scoring and filtering. }We score each candidate task $\hat{t} \in \widetilde{\mathcal{T}}^{(s)}$ (including both original tasks in $\mathcal{T}^{(s)}$ and augmented tasks in $\mathcal{A}(\mathcal{T}^{(s)})$) along three dimensions:
\begin{itemize}[leftmargin=1.2em]
  \item \textbf{Question quality} $S_{\text{qq}}(\hat{t})$  clarity, completeness, and effectiveness as a realistic user query.
  \item \textbf{Scenario realism} $S_{\text{sr}}(\hat{t})$  authenticity and plausibility of the described scenario.
  \item \textbf{Tool-use necessity} $S_{\text{tn}}(\hat{t})$  whether tool use is necessarily required and appropriately selected (i.e., the task is not trivially solvable without tools).
\end{itemize}
We retain a candidate only if it passes all thresholds:
\begin{equation}
S_{\text{qq}}(\hat{t}) \ge \theta_{\text{qq}},\quad
S_{\text{sr}}(\hat{t}) \ge \theta_{\text{sr}},\quad
S_{\text{tn}}(\hat{t}) \ge \theta_{\text{tn}}.
\label{eq:task_filter}
\end{equation}
Candidates failing Eq.~\eqref{eq:task_filter} are discarded.

\subsubsection{Trajectory Collection via Multi-turn Interaction}
\label{subsec:trajectory_collection}

We use \texttt{Qwen-Agent} \citep{qwen-agent} to handle the tool-calling loop.

\paragraph{Tool composition and hybrid execution.}
Our tool pool consists of two categories:
\begin{itemize}[leftmargin=1.2em]
  \item \textbf{Deployed MCP servers }
  Tool calls are executed directly at runtime, and returned outputs are logged as environment feedback.

  \item \textbf{Doc-only MCP servers }
  We employ a stateful tool-response emulation module to synthesize plausible outputs.
  The emulator retains session-level invocation histories and synthesized outputs to ensure coherent multi-turn interactions.
  To approximate real-world unreliability, we additionally inject tool failures into the emulation process, causing emulated calls to fail with a probability of $20\%$ (e.g., due to timeouts or unreachable calls).

\end{itemize}

\subsubsection{Reward Modeling}
\label{subsec:reward_deferred}

To enable high-quality supervised fine-tuning for tool-augmented language models, we design an automated trajectory quality assessment pipeline without human annotation. We define a trajectory as an ordered sequence 
\begin{equation}
\tau = \{m_0, m_1, \ldots, m_{k-1}\},
\end{equation}
where $k$ is the total number of messages in the trajectory, $m_0$ is the system prompt, $m_1$ denotes the user query $q$, and $m_{i \ge 2}$ represents the subsequent interaction turns, including assistant responses, tool invocations, and environment responses.

\paragraph{Query Understanding and Planning.}
The initial assistant response $m_2$ both interprets the user query and proposes an initial plan that guides subsequent tool use and interaction. We therefore evaluate these two aspects separately (while using the same trajectory-level input), so that failures due to misunderstanding can be distinguished from failures due to infeasible planning. In both cases, we exclude the system prompt from the evaluator input:
\begin{align}
\text{QU}(\tau) &= \mathcal{J}_{\text{understand}}\big(\tau \setminus \{m_0\}\big) \in \{0,0.5,1\}, \\
\text{QP}(\tau) &= \mathcal{J}_{\text{plan}}\big(\tau \setminus \{m_0\}\big) \in \{0,0.5,1\}.
\end{align}
A score of $1$ indicates correct understanding (or a complete and executable plan), $0.5$ corresponds to partial understanding (or a partially feasible plan), and $0$ indicates misunderstanding (or an infeasible/incorrect plan).

\paragraph{Tool Response Understanding and Planning.}
We define two trajectory-level metrics:
\begin{itemize}[leftmargin=*,nosep]
    \item \textbf{Tool-response Context Understanding (TCU) } a trajectory-level score that measures whether each tool-call round reflects correct understanding of the immediately preceding tool response.
    \item \textbf{Tool-response Context-conditioned Planning (TCP) } a trajectory-level score that measures whether each tool-call round’s plan/tool invocation(s) correctly incorporate that tool response.
\end{itemize}
If a turn contains multiple tool calls, we merge them into one grouped round.
Let $\{u_j\}_{j=1}^{J}$ denote the grouped tool-call rounds in temporal order, and let $t_j$ be the index of the assistant message that issues $u_j$.
Since $u_1$ has no preceding tool response, we score from $j=2$ using the same history-plus-current-round input:
\begin{equation}
c_j \triangleq \{m_1,\ldots,m_{t_j}\}, \qquad j=2,\ldots,J.
\end{equation}
We compute the trajectory-level scores by averaging per-round judgments with inputs $(c_j,u_j)$:
\begin{align}
\mathrm{TCU}(\tau) &= \frac{1}{J-1}\sum_{j=2}^{J} \mathcal{J}_{\text{understand}}(c_j, u_j), \\
\mathrm{TCP}(\tau) &= \frac{1}{J-1}\sum_{j=2}^{J} \mathcal{J}_{\text{plan}}(c_j, u_j).
\end{align}

\paragraph{Tool Call Status.}
Let $n$ denote the total number of tool calls in trajectory $\tau$. 
For the $i$-th tool call, we assign a binary indicator $S_i \in \{0,1\}$, where $S_i=1$ if the call succeeds (i.e., returns a valid response) and $S_i=0$ otherwise. 
The trajectory-level tool-call status score is computed as the mean success rate:
\begin{equation}
\mathrm{TCS}(\tau) = \frac{1}{n}\sum_{i=1}^{n} S_i.
\end{equation}

\paragraph{Tool Conciseness.}
For the $i$-th tool call, we assign a binary indicator $\mathrm{TC}_i \in \{0,1\}$, where $\mathrm{TC}_i=1$ if the call is necessary and non-redundant given the task and prior context, and $0$ otherwise. 
We report the trajectory-level conciseness score as:
\begin{equation}
\mathrm{TC}(\tau) = \frac{1}{n}\sum_{i=1}^{n} \mathrm{TC}_i.
\end{equation}
A higher score indicates efficient tool usage without redundant calls, while lower scores indicate unnecessary or inefficient invocations.

\paragraph{Final Answer Quality.}
The final answer quality evaluates whether the last assistant message $m_{k-1}$ is both (i) semantically aligned with the original task specification and (ii) faithful to the trajectory content. Specifically, we measure semantic correlation between the user prompt and the final answer, and assess faithful summarization over the trajectory excluding the system prompt:
\begin{equation}
\text{FA}(\tau) = \frac{\text{Corr}(m_1, m_{k-1}) + \text{Summ}(\tau \setminus \{m_0\})}{2}.
\end{equation}

The above modules produce a set of seven trajectory-level scores. We aggregate them into a single scalar reward by taking the arithmetic mean across the seven dimensions.


\subsection{Automated Verifiable Environment Synthesis}

\begin{figure}[h!]
    \centering
    \includegraphics[width=1.0\linewidth]{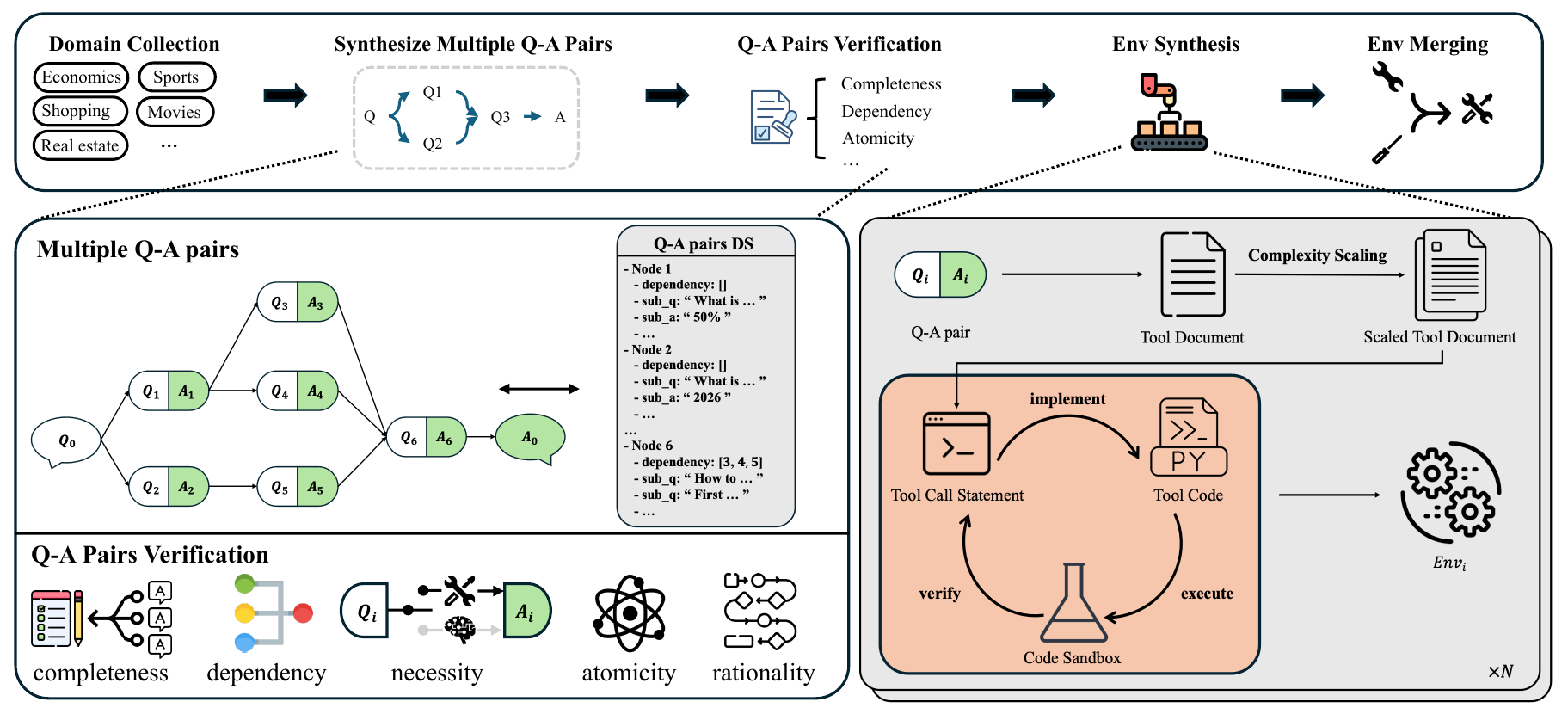}
    \caption{Overview of the QA-Based Environment Synthesis Framework.}
    \label{fig:Env}
\end{figure}

Overall, our environment synthesis pipeline consists of four major stages: \textbf{Q--A Instance Synthesis}, \textbf{Quality Validation}, \textbf{Environment Synthesis}, and \textbf{Sub-Environment Merging}, as depicted in Figure~\ref{fig:Env}.

\subsubsection{Q--A Instance Synthesis as Semantic Topology Extraction}
We argue that an LLM's agent capability depends on its ability to learn the latent planning and tool-use patterns underlying human cognition---selecting actions, updating task state from tool feedback, and replanning over multiple turns. 

Unlike static path-supervised tool chains,we model multi-turn tool use as navigation over a latent semantic topology, verify only sub-tasks attainment, and optimize a composite reward that optimizes for success while penalizing interaction cost—rather than prescribing a fixed tool chain.

We formalize each instance as a \textbf{main question} $q_0$ together with its \textbf{main answer} $a_0$. During the solution process, the model often needs to resolve a set of intermediate sub-tasks. We explicitly represent these intermediate steps as a collection of \textbf{sub-questions} and \textbf{sub-answers}:
\begin{equation}
\mathcal{S}=\{(q_i,a_i)\}_{i=1}^{m},
\end{equation}
where each pair $(q_i,a_i)$ corresponds to a necessary or helpful intermediate step for deriving $a_0$, and $m$ denotes the total number of such intermediate pairs. We model the derivation of the final answer as aggregating the sub-answers according to their dependency graph:
\begin{equation}
a_0 = \Phi\big(\{a_i\}_{i=1}^{m}, \mathcal{G}\big),
\end{equation}
where $\mathcal{G}$ denotes the dependency structure among sub-questions (e.g., an ordered chain or a DAG), and $\Phi(\cdot)$ denotes an aggregation procedure that combines sub-answers following $\mathcal{G}$. By jointly generating $(q_0,a_0)$ and the intermediate steps $\mathcal{S}$, we obtain an explicit and verifiable representation of the solution process.

\paragraph{Synthesis Mode.}

Each instance is synthesized conditioned on a domain-specific knowledge source $\mathcal{K}$ (e.g., a text corpus) and a complexity constraint $H$ (hop budget). The synthesis process follows two modes: 
\begin{itemize}[leftmargin=1.2em]
    \item \textbf{Question-Conditional Generation } If a specific main question $q_0$ is provided, the module decomposes it into a sub-QA set $\mathcal{S}$ and derives $a_0$ based on $\mathcal{K}$.
    \item \textbf{Unconditional Generation } If $q_0$ is not provided, the module first generates a candidate question $q_0$ grounded in $\mathcal{K}$ that requires approximately $H$ reasoning hops, and subsequently produces the corresponding $(q_0, a_0)$ and $\mathcal{S}$.
\end{itemize}

\subsubsection{Quality Validation}

We observe that a subset of synthesized Q--A instances contains intermediate sub-questions that do not require tool invocation. These sub-questions correspond to purely linguistic operations, such as evaluation, summarization, recommendation, advice, ranking, matching, and format transformation. Such steps cannot be grounded in executable tools and thus disrupt the continuity of the tool-use chain, preventing the construction of a fully verifiable agent environment.

Specifically, we allow sub-questions that do not require tool invocation only at leaf nodes and prohibit them at non-leaf nodes. Leaf nodes typically correspond to final linguistic aggregation or answer formulation steps, which do not require external tools and do not introduce downstream dependencies.

We first filter out samples whose intermediate Q–A pairs do not require tool usage using LLM. After filtering, we assign a quality score to each remaining Q--A pair along four complementary dimensions.

\paragraph{Dependency Consistency. }We formalize a decomposed Q--A instance as a set of $m$ sub-questions:
\begin{equation}
\tau = \{(q_1, a_1, d_1), (q_2, a_2, d_2), \ldots, (q_m, a_m, d_m)\},
\end{equation}
where $q_i$ and $a_i$ denote the $i$-th sub-question and its corresponding answer, and $d_i$ represents the dependency set of sub-question $q_i$, specifying which preceding sub-questions and their corresponding answers it depends on.

To assess dependency consistency, we leverage LLM to verify each dependency set $d_i$ by judging whether all listed dependencies are semantically and logically necessary for answering $q_i$. For each sub-question, the dependency score is defined as a binary indicator:
\begin{equation}
\mathrm{DC}_i = \begin{cases}
1, & \text{if all dependencies in } d_i \text{ are correct}, \\
0, & \text{otherwise.}
\end{cases}
\end{equation}

The overall dependency consistency score for a Q--A instance is then computed as the average over all sub-questions:
\begin{equation}
\mathrm{DC}(\tau) = \frac{1}{m} \sum_{i=1}^{m} \mathrm{DC}_i.
\end{equation}

\paragraph{Sub-Question Atomicity.}

Sub-question atomicity evaluates whether each sub-question corresponds to an indivisible unit that cannot be further decomposed. Given a decomposed Q--A instance $\tau = \{(q_i, a_i, d_i)\}_{i=1}^{m}$, each sub-question is evaluated by LLM to determine whether it is atomic. An atomic sub-question receives a score of 1; otherwise, it receives a score of 0:
\begin{equation}
\mathrm{SA}_i = \begin{cases}
1, & \text{if } q_i \text{ is atomic}, \\
0, & \text{otherwise.}
\end{cases}
\end{equation}

The overall atomicity score is computed as the average over all sub-questions:
\begin{equation}
\mathrm{SA}(\tau) = \frac{1}{m} \sum_{i=1}^{m} \mathrm{SA}_i.
\end{equation}

\paragraph{Sequential Rationality.}

When synthesizing Q--A instances, we explicitly specify the expected number of reasoning hops. We observe that, in some cases, the language model introduces logically inconsistent or unnatural transitions solely to satisfy the prescribed hop count, resulting in irrational execution orders within the Q--A instance. To address this issue, we design a sequential rationality checking module to assess whether the ordering of sub-questions is logically valid.

Formally, given a decomposed Q--A instance $\tau = \{(q_i, a_i, d_i)\}_{i=1}^{m}$, we evaluate sequential rationality based on the dependency sets $d_i$. A sub-question is considered rational if each sub-question $q_i$ is executed only after all its dependencies in $d_i$ have been satisfied, and no superfluous intermediate steps are introduced. For each sub-question, the sequential rationality score is defined as a binary indicator:
\begin{equation}
\mathrm{SR}_i = \begin{cases}
1, & \text{if the execution order implied by } d_i \text{ is rational}, \\
0, & \text{otherwise.}
\end{cases}
\end{equation}

The overall sequential rationality score for a Q--A instance is computed as the average over all sub-questions:
\begin{equation}
\mathrm{SR}(\tau) = \frac{1}{m} \sum_{i=1}^{m} \mathrm{SR}_i.
\end{equation}

\paragraph{Task Completeness.}

To verify that the decomposition is logically consistent, we evaluate task completeness by checking whether the set of sub-questions is sufficient to solve the original task.
Given a decomposed Q--A instance $\tau = \{(q_i, a_i, d_i)\}_{i=1}^{m}$, we define an instance-level binary score:
\begin{equation}
\mathrm{TC}(\tau) =
\begin{cases}
1, & \text{if } \{(q_i, a_i)\}_{i=1}^{m} \text{ is sufficient to solve the main question}\, q_0, \\
0, & \text{otherwise.}
\end{cases}
\end{equation}

We combine the four quality dimensions--dependency consistency, sub-question atomicity, sequential rationality, and task completeness--to obtain an overall quality score for each decomposed Q--A instance. 
Formally, for a decomposition instance $\tau$, we define $\mathrm{QS}(\tau)$ as its aggregated quality score, computed by averaging the four dimension-specific scores:
\begin{equation}
\mathrm{QS}(\tau) = \frac{1}{4} \Bigl( \mathrm{DC}(\tau) + \mathrm{SA}(\tau) + \mathrm{SR}(\tau) + \mathrm{TC}(\tau) \Bigr).
\end{equation}

\subsubsection{Environment Synthesis}

Following quality validation, we obtain a filtered dataset \( \mathcal{D}_{\text{filtered}} \). As illustrated in Figure~\ref{fig:Env}, we synthesize an independent environment for each Q--A instance. Formally, each instance is represented as a decomposed execution trace \( \tau = \{(q_i, a_i, d_i)\}_{i=1}^{m} \), where each triplet \( (q_i, a_i, d_i) \) denotes a sub-task node consisting of a sub-question \( q_i \), its ground-truth answer \( a_i \), and the associated dependency set \( d_i \).

We skip leaf nodes and synthesize sub-environments only for the remaining ones, treating each sub-task \( (q_i, a_i, d_i) \) as an independent sub-environment.

\paragraph{Tool Specification Synthesis and Complexity Scaling.}
We first feed \( (q_i, a_i, d_i) \) into LLM to generate a tool specification document that describes the tool's functionality, input parameters, and expected outputs. To improve expressiveness and better support diverse tool-invocation patterns, we further augment the generated specification by scaling its complexity, e.g., expanding parameter lists and enriching parameter value spaces through additional arguments or extended enumerated ranges.

\paragraph{Tool Implementation and Sandbox Verification.}
Conditioned on \( q_i \) and the augmented tool document, we then generate a tool invocation statement. Subsequently, we use \( (q_i, a_i) \) together with the tool specification and invocation statement to synthesize a Python-based tool implementation. The generated code is executed in a sandboxed environment for validation, 
where a sub-environment is considered successful if the execution result contains the target answer \(a_i\). Otherwise, we restart the process from the tool invocation statement generation step and repeat for a fixed number of attempts.

After all sub-task nodes in \( \tau \) have been successfully synthesized and validated, we aggregate the resulting sub-environments into a unified collection. This collection constitutes a complete, standalone, and executable environment for the original Q--A instance, enabling deterministic execution and verification across the entire decomposed trace.

\subsubsection{Sub-Environment Merging}

To avoid action space inflation caused by functionally equivalent sub-questions, we perform \textbf{intra-instance sub-environment merging} to remove such redundancies. We carry out sub-environment merging in two stages.

\paragraph{Homogeneous Sub-Question Identification.}
Given a Q--A decomposed trace $\tau = \{(q_i, a_i, d_i)\}_{i=1}^{m}$, we use LLM to identify homogeneous sub-questions that share the same functional intent but differ in their parameter values (e.g., weather queries for different cities). Based on this classification, we group sub-questions into $n$ homogeneous sets($n \leq m$) and obtain a merged representation:
\begin{equation}
\tau' = \{ s_1, s_2, \ldots, s_n \},
\end{equation}
where each $s_j$ denotes a set of triplets corresponding to homogeneous sub-questions.

\paragraph{Database Expansion.}
For each homogeneous set $s_j$, we randomly select one triplet as the base instance and treat its synthesized tool implementation as the initial sub-environment. We then iteratively insert the remaining triplets in $s_j$ into the Python implementation by extending the underlying data structures, while the corresponding invocation statements are generated by an LLM. After each insertion, we execute all existing invocation statements in a sandboxed environment to verify that the tool can still return correct answers for all associated sub-questions.

After completing this procedure for all $s_j$, we obtain a merged set of sub-environments in which functionally equivalent tools are consolidated into a single implementation while preserving correctness for all original triplets in $\tau$.

\section{Training and Evaluation of Tool Agents}

This section describes how we train and evaluate ASTRA. \textbf{First}, we summarize the key improvements in our training infrastructure. \textbf{Second}, we detail the two-stage training settings, covering both SFT and RL. \textbf{Finally}, we introduce the benchmarks and report evaluation results.

\subsection{Infrastructure}

We summarize the key infrastructure improvements for both SFT and RL that enable efficient training and stable online optimization in ASTRA.

\paragraph{SFT Infrastructure.}

We perform SFT using the HuggingFace Transformers library\footnote{https://github.com/huggingface/transformers}. To characterize tool-use learning dynamics, we save checkpoints at a high frequency for fine-grained tracking. Each full checkpoint in Transformers typically bundles both model parameters and the complete training state (e.g., optimizer and scheduler states), which substantially increases storage overhead under frequent saving. To mitigate the I/O overhead that can slow training and the storage overhead incurred by frequent checkpointing , we modify the checkpointing pipeline to decouple parameter snapshots from training-state serialization: we persist model weights at high frequency, while retaining training-state checkpoints only for the most recent 1--2 saves. This design preserves fine-grained observability for analysis and ablations, while keeping storage requirements practical at scale.

\paragraph{RL Infrastructure.}

Our reinforcement learning pipeline is implemented using verl\footnote{https://github.com/volcengine/verl}. We frame reinforcement learning as interactive tool-use over a collection of instance-specific, fully isolated simulators: each training instance is paired with an independent environment, and no state or information is shared across instances. Unlike prior approaches that roll out a single trajectory and apply updates only at individual tool-invocation nodes, we adopt an \textbf{online, multi-turn agentic reinforcement learning paradigm}.

\begin{figure}[h!]
    \centering
    \includegraphics[width=1.0\linewidth]{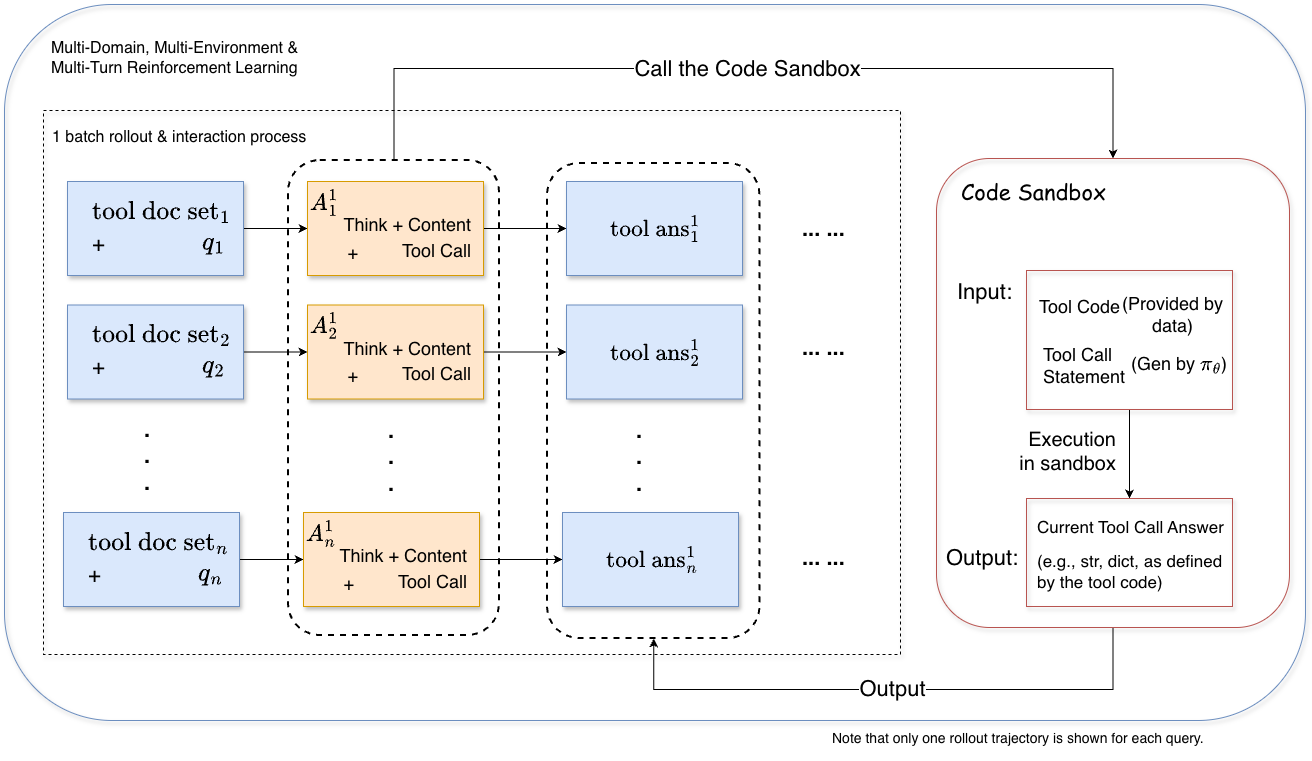}
    \caption{Rollout Procedure in Reinforcement Learning.}
    \label{fig:RL}
\end{figure}

As illustrated in Figure~\ref{fig:RL}, at each interaction step the policy model generates a tool invocation statement. This statement, together with the corresponding tool implementation code, is passed into a code sandbox. The sandbox executes the tool call and returns the tool output produced under the current environment state. The returned result is then fed back to the model as an observation, enabling the agent to condition subsequent decisions on the accumulated interaction history.

For each data instance, the multi-turn interaction terminates when any of the following conditions is met:

\begin{itemize}

\item The interaction reaches a predefined maximum number of turns or a maximum sequence length

\item The model stops issuing tool calls, i.e., no further tool invocation is generated

\end{itemize}

Under these termination criteria, the collected trajectory—comprising inputs, observations, tool calls, tool outputs, and rewards—is used directly for online policy optimization. This formulation allows the model to \textbf{learn long-horizon decision-making strategies} over tool-augmented environments, rather than optimizing isolated single-step actions.

For policy optimization, we build on the GRPO~\citep{shao2024deepseekmathpushinglimitsmathematical} objective (Equation~\ref{eq_grpo}). In our implementation, we omit both the KL-divergence regularizer and the entropy bonus for simplicity and empirical stability. However, under this simplified objective, if all samples within a group $G$ receive identical rewards, the resulting advantage estimates collapse to zero, yielding no gradient signal and effectively reducing the number of learning-active samples per nominal batch. This mismatch can introduce training instability.

\begin{equation}
\begin{aligned}
\mathcal{J}_{\mathrm{GRPO}}(\theta)
&=
\mathbb{E}\!\left[
q \sim P(Q),\ \{o_i\}_{i=1}^{G} \sim \pi_{\theta_{\mathrm{old}}}(\cdot \mid q)
\right] \\[6pt]
&\quad
\frac{1}{G}
\sum_{i=1}^{G}
\frac{1}{|o_i|}
\sum_{t=1}^{|o_i|}
\Bigl\{
\min \left(
\frac{\pi_\theta(o_{i,t} \mid q, o_{i,<t})}
     {\pi_{\theta_{\mathrm{old}}}(o_{i,t} \mid q, o_{i,<t})}
\hat{A}_{i,t},
\;
\operatorname{clip}\!\left(
\frac{\pi_\theta(o_{i,t} \mid q, o_{i,<t})}
     {\pi_{\theta_{\mathrm{old}}}(o_{i,t} \mid q, o_{i,<t})},
1-\epsilon,\ 1+\epsilon
\right)
\hat{A}_{i,t}
\right) \\
&\qquad
-\,
\beta D_{\mathrm{KL}}\!\left(
\pi_\theta \,\|\, \pi_{\mathrm{ref}}
\right)
\Bigr  \}.
\label{eq_grpo}
\end{aligned}
\end{equation}

\begin{figure}[h!]
    \centering
    \includegraphics[width=1.0\linewidth]{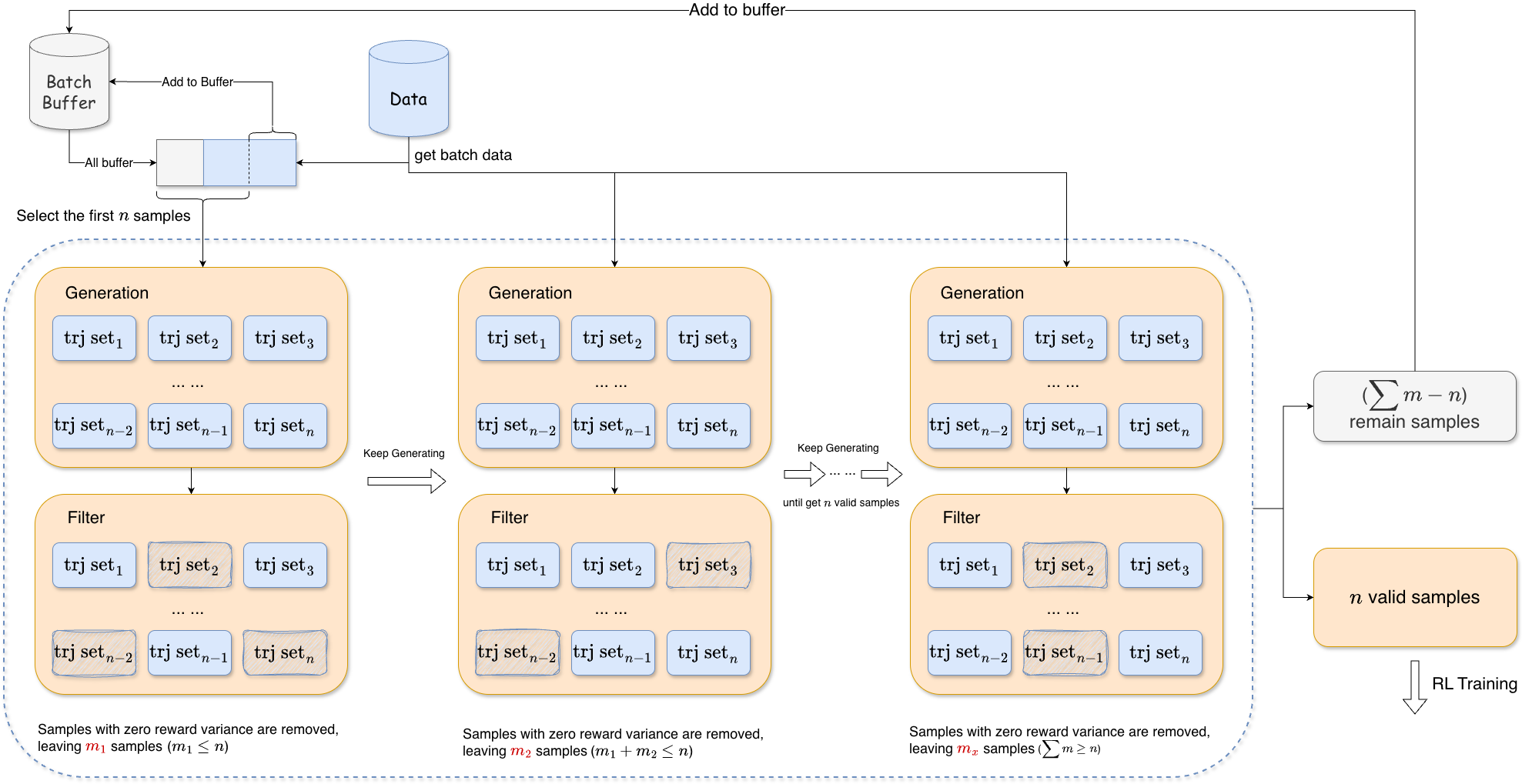}
    \caption{One-Step Adaptive Batch Filling.}
    \label{fig:RL_cb}
\end{figure}

To address this problem, we adopt \textbf{Adaptive Batch Filling}, a simple yet effective batching strategy illustrated in Figure~\ref{fig:RL_cb}. Let $n$ denote the target batch size. Here, we call a rollout valid if it yields a non-zero learning signal--operationally, if the reward variance within its GRPO group is non-degenerate (e.g., $\operatorname{Std}(R)>\delta$, cf. Equation~\ref{eq_grpo_ours}). We maintain a data buffer that is initially empty and always satisfies $|buffer| < n$. Before each rollout, we retrieve all samples currently stored in the buffer and concatenate them with newly batch samples.

If the concatenated set contains more than $ n $ valid samples, we select the first $ n $ samples to form the current training batch, while the remaining valid samples are placed back into the buffer. Rollout generation continues until the number of valid samples is greater than or equal to $n$, ensuring that each optimization step is performed with a full batch of effective training data.

\subsection{Training Settings}

\subsubsection{SFT Settings}
We perform SFT on two models, \texttt{Qwen3-14B} and \texttt{Qwen3-32B}~\citep{qwen3technicalreport}. All models are trained for two epochs with a maximum sequence length of 20k tokens without packing. We use a batch size of 32 for all SFT experiments.

The learning rate is set to $5\times10^{-6}$ for \texttt{Qwen3-14B} and $2\times10^{-6}$ for \texttt{Qwen3-32B}. In both cases, we adopt a cosine learning rate schedule with a warmup ratio of 5\% of the total training steps.

To support long-context training efficiently, we employ Context Parallelism (also referred to as Sequence Parallelism). Specifically, we use a context parallel degree of CP=2 when training \texttt{Qwen3-14B}, and CP=4 when training \texttt{Qwen3-32B}.

\subsubsection{RL Settings}
\paragraph{Irrelevant Tool Mixing.} \label{sec:irrelevant_tool_mixing}

To improve robustness in tool selection across diverse tools, we augment each training instance with a controlled number of task-irrelevant tools, drawn from multiple semantic similarity bands. This expands the tool inventory beyond the minimal set required to solve the task, encouraging the model to discriminate truly relevant tools rather than overfitting to a fixed or overly clean tool list.

Let $\mathcal{T}$ denote the global tool pool obtained from all environments. We first remove duplicates by exact-match deduplication on tool names, yielding a unique set $\mathcal{T}_{\mathrm{uniq}}=\{\tau_1,\dots,\tau_M\}$. Each tool $\tau$ is associated with an OpenAI-standard tool documentation string $d(\tau)$, which typically includes a concise tool description and argument description that specify each tool's expected usage. We embed $d(\tau)$ using \texttt{Qwen3-Embedding-8B}~\citep{qwen3embedding}:
\begin{equation}
\mathbf{e}_\tau = f\big(d(\tau)\big) \in \mathbb{R}^D, \qquad \tau \in \mathcal{T}_{\mathrm{uniq}}.
\end{equation}
Based on these embeddings, we compute a cosine similarity matrix over tools,
\begin{equation}
S_{ij} = \mathrm{cos}\left(\mathbf{e}_{\tau_i},\mathbf{e}_{\tau_j}\right), \qquad 1\le i,j\le M.
\end{equation}

For a training instance $x$, the environment exposes an instance-specific tool set
$\mathcal{T}(x)\subseteq \mathcal{T}_{\mathrm{uniq}}$.
For each in-scope tool $\tau_i\in\mathcal{T}(x)$, we normalize its similarity to every other tool:
\begin{equation}
\begin{aligned}
{} & S_i^{\min} := \min_{j\neq i} S_{ij}, \\
{} & S_i^{\max} := \max_{j\neq i} S_{ij}, \\
{} & \widetilde{S}_{ij} := \frac{S_{ij}-S_i^{\min}}{S_i^{\max}-S_i^{\min}} \in [0,1].
\end{aligned}
\end{equation}

To avoid near-duplicate tools that could destabilize training, we exclude same-domain candidates when forming
similarity-based pools for $\tau_i$.

Using fixed thresholds on $\widetilde{S}_{ij}$, we partition the remaining candidates into three semantic similarity bands:
\begin{equation}
\begin{aligned}
\mathcal{B}^{\mathrm{high}}_i(x) &:= \{\tau_j : \widetilde{S}_{ij} > 0.85\}, \\
\mathcal{B}^{\mathrm{med}}_i(x)  &:= \{\tau_j : 0.4 \le \widetilde{S}_{ij} \le 0.85\}, \\
\mathcal{B}^{\mathrm{low}}_i(x)
&:= \{\tau_j : \widetilde{S}_{ij} < 0.4\}.
\end{aligned}
\end{equation}

For each instance $x$ and similarity band $b\in\{\mathrm{high},\mathrm{med},\mathrm{low}\}$,
let $\mathcal{P}^{b}(x)$ denote the instance-level candidate set of tools in band $b$,
obtained by aggregating the per-tool candidate sets $\{\mathcal{B}^{b}_i(x)\}$ over all in-scope tools:
\begin{equation}
\mathcal{P}^{b}(x) := \bigcup_{\tau_i\in\mathcal{T}(x)} \mathcal{B}^{b}_i(x),
\qquad b\in\{\mathrm{high},\mathrm{med},\mathrm{low}\}.
\end{equation}
We then uniformly sample up to $K$ tools from each of $\mathcal{P}^{high}(x)$, $\mathcal{P}^{med}(x)$, and
$\mathcal{P}^{low}(x)$; the sampled tools are used to augment the tool list presented to the model for instance $x$.

\paragraph{Reward Design.}

As described earlier, each data instance is associated with a set of sub-tasks to be solved, which we formalize as a job consisting of multiple question–answer pairs:

\begin{equation}
job = \{ (q_1, a_1), (q_2, a_2), \dots , (q_n, a_n)\}
\end{equation}

where each pair $(q_i, a_i)$ denotes a sub-question and its corresponding ground-truth answer.

Given a policy $\pi_\theta$, suppose that during a multi-turn interaction the agent invokes tools $c$ times and successfully solves $\hat{n}$ sub-tasks. We evaluate the resulting trajectory using an F1-style reward that jointly accounts for \textbf{task completion and interaction efficiency}. Specifically, we define

\begin{equation}
r = \frac{\hat{n}}{n}, p = \frac{\hat{n}}{c + \epsilon}
\end{equation}

where $r$ measures sub-task recall, i.e., the fraction of required sub-tasks that are successfully solved, and $p$ measures precision with respect to tool usage, i.e., the effectiveness of each tool invocation.

The final trajectory-level reward is then computed as the harmonic mean of $p$ and $r$:

\begin{equation}
reward = \frac{2pr}{p + r}
\end{equation}

This reward design explicitly encourages the agent to solve as many sub-tasks as possible while minimizing redundant or unnecessary tool calls. By operating at the trajectory level, the reward provides a dense yet structured training signal for online multi-turn reinforcement learning, promoting long-horizon planning and efficient tool utilization in executable, verifiable environments.

\begin{equation}
\begin{aligned}
&\mathcal{J}_{\mathrm{GRPO}}'(\theta)
 \\ &= 
\mathbb{E}_{(q,a)\sim\mathcal{D},\ \{o_i\}_{i=1}^{G}\sim\pi_{\theta_{\mathrm{old}}}(\cdot\mid q)}
\!\left[
\,\cdot\ \Bigm|\ 
\textcolor{red}{\operatorname{Std}\!\big(R(q,\{o_i\})\big)> \delta}
\right]
\\[6pt]
&\quad
\left[
\frac{1}{\textcolor{red}{\sum_{i=1}^{G} |o_i|}}
\sum_{i=1}^{G}
\sum_{t=1}^{|o_i|}
\min \left(
\frac{\pi_\theta(o_{i,t}\mid q,o_{i,<t})}
     {\pi_{\theta_{\mathrm{old}}}(o_{i,t}\mid q,o_{i,<t})}
\hat{A}_{i,t},
\;
\operatorname{clip}\!\left(
\frac{\pi_\theta(o_{i,t}\mid q,o_{i,<t})}
     {\pi_{\theta_{\mathrm{old}}}(o_{i,t}\mid q,o_{i,<t})},
1-\epsilon,
1+\epsilon
\right)
\hat{A}_{i,t}
\right)
\right]
\label{eq_grpo_ours}
\end{aligned}
\end{equation}

Following prior work~\citep{yu2025dapoopensourcellmreinforcement}, instead of averaging the token loss at the sequence level, we adopt a batch-level token loss averaging strategy. Empirically, we find that combining batch-level token loss averaging with Adaptive Batch Filling leads to stable training dynamics and consistent performance improvements throughout training. \textbf{The final reinforcement learning objective is defined in Equation~\ref{eq_grpo_ours}}.

\paragraph{Training configurations.} We set both the batch size and the mini-batch size to 256. This choice corresponds to a strictly online learning setting, where each optimization step is performed on collected trajectories without replaying past samples. The learning rate is fixed to $2\times10^{-6}$ across all reinforcement learning experiments.

We adopt long-context settings to support multi-turn agent interactions. Specifically, the maximum prompt length is set to 25,600 tokens, and the maximum response length is set to 49,152 tokens. For each trajectory, we allow up to 32 turns for both the user and the assistant, enabling the agent to handle long-horizon, multi-step tool-use scenarios during training.

\subsection{Evaluation}
\subsubsection{Benchmarks}

We primarily evaluate our models on agentic multi-turn tool use.
In addition, we include an evaluation on non-agentic complex reasoning to assess core reasoning competence.

\paragraph{Agentic benchmarks.}
We use three widely adopted interactive benchmarks: BFCL-v3 Multi-Turn (abbreviated as BFCL-MT)~\citep{patil2025bfcl}, $\tau^2$-Bench~\citep{barres2025tau2}, and ACEBench~\citep{chen2025acebench}. Each benchmark provides domain-specific environments equipped with tools, requiring multi-step and multi-turn interaction and integrating tool outputs into subsequent decisions. $\tau^2$-Bench and ACEBench further include a user simulator, stressing robustness under interactive user feedback. For $\tau^2$-Bench, we exclude the airline subset due to concerns about lower-quality ground-truth grading noted by prior reports~\citep{openai2025gpt52}.

\paragraph{Non-agentic benchmarks.}
To verify that our method enhances agentic behavior without sacrificing core logical reasoning, we additionally evaluate on AIME2024~\citep{maa_aime_2024} and AIME2025~\citep{ye2025aimepreview}, which focus on mathematical problem solving.

\subsubsection{Evaluation Methodology}

\begin{table*}[t]
\centering
\caption{\textbf{Agentic benchmark results.} Performance on BFCL-MT, $\tau^2$-Bench, and ACEBench across multiple model scales, covering closed-source, open-source, and our models.}
\label{tab:bfcl_tau2_acebench_results}
\footnotesize
\setlength{\tabcolsep}{2.2pt}
\renewcommand{\arraystretch}{1.15}
\resizebox{\textwidth}{!}{%
\begin{tabular}{@{}l rrrrr @{\hskip 8pt} r @{\hskip 2.5pt} c @{\hskip 2pt} c @{\hskip 8pt} rrr@{}}

\toprule
& \multicolumn{5}{c}{BFCL-MT}
& \multicolumn{3}{c}{$\tau^2$-Bench}
& \multicolumn{3}{c}{ACEBench} \\

\cmidrule(lr){2-6}\cmidrule(lr){7-9}\cmidrule(lr){10-12}
Model
& Base & \shortstack{Missing\\Func} & \shortstack{Missing\\Param} & \shortstack{Long\\Context} & Overall
& Retail & Telecom & Overall
& \shortstack{Multi\\Turn} & \shortstack{Multi\\Step} & Overall \\
\midrule

\multicolumn{12}{@{}l}{\textbf{Closed-source}} \\
Claude-Opus-4.5~\citep{anthropic2025opus45}   & 81.00 & 64.00 & 58.00 & 70.50 & 68.38 & 80.88 & 90.70 & 85.79 & 64.17 & 100.00 & 82.09 \\
Gemini-3-Pro~\citep{google2025gemini3pro}      & 69.00 & 63.00 & 56.50 & 64.00 & 63.13 & 77.72 & 89.65 & 83.69 & 52.09    & 100.00      & 76.05    \\
Claude-Sonnet-4.5~\citep{anthropic2025sonnet45} & 69.00 & 65.00 & 52.50 & 59.00 & 61.38 & 77.19 & 75.96 & 76.58 & 65.83 & 94.38   & 80.11 \\
Claude-Haiku-4.5~\citep{anthropic2025haiku45}  & 63.50 & 42.50 & 52.50 & 56.00 & 53.63 & 69.12 & 37.19 & 53.16 & 64.17 & 88.75   & 76.46 \\
GPT-4.1~\citep{openai2025gpt41}                    & 47.50 & 32.50 & 32.50 & 43.00 & 38.88 & 74.00 & 34.00 & 54.00 & 66.67 & 95.00   & 80.84 \\
Gemini-2.5-Pro~\citep{google2025gemini25pro}             & 28.50 & 35.00 & 30.00 & 27.00 & 30.12 & 71.26 & 37.89 & 54.58 & 97.50    & 40.00      & 68.75    \\
\midrule

\multicolumn{12}{@{}l}{\textbf{Open-source}} \\
Kimi-K2-Instruct~\citep{kimiteam2025kimik2openagentic}  & 62.00 & 41.00 & 44.50 & 55.00 & 50.63 & 68.64    & 70.39    & 69.52    & 69.17 & 92.50 & 80.84 \\
GLM-4.6~\citep{5team2025glm45agenticreasoningcoding}                     & 74.50 & 68.00 & 63.00 & 66.50 & 68.00 & 78.95    & 60.31    & 69.63    & 72.50 & 87.50 & 80.00 \\
LoopTool-32B~\citep{zhang2025looptoolclosingdatatrainingloop}                & 66.50 & 58.00 & 44.50 & 62.00 & 57.75 & 72.15 & 48.46 & 60.31 & 57.50 & 60.00 & 58.75 \\

Qwen3-32B~\citep{qwen3technicalreport}                   & 59.00 & 47.50 & 40.50 & 51.50 & 49.63 & 64.70 & 34.70 & 49.70 & 55.83 & 63.75 & 59.79 \\
Qwen3-14B~\citep{qwen3technicalreport}                   & 54.00 & 39.50 & 39.00 & 45.50 & 44.50 & 55.00 & 37.10 & 46.05 & 45.83 & 57.50 & 51.67 \\

\midrule

\multicolumn{12}{@{}l}{\textbf{Astra (ours)}} \\
Astra-14B-thinking-v1       & 67.00 & 56.00 & 48.50 & 61.00 & 58.13 & 68.00 & 47.37 & 57.69 & 54.17 & 83.75 & 68.96 \\
Astra-32B-thinking-v1       & 76.50 & 65.50 & 48.50 & 66.50 & 64.25 & 75.20 & 52.19 & 63.70 & 60.00 & 83.75 & 71.88 \\

\bottomrule
\end{tabular}%
}
\end{table*}

\paragraph{Agentic Evaluation.}
All experiments are executed with vLLM~\citep{kwon2023efficient} as the inference engine to ensure consistent serving and decoding behavior across benchmarks.
To faithfully reflect agentic capability, we evaluate all tool-use benchmarks under the function-calling paradigm.
Benchmark-specific settings are as follows:
\begin{itemize}[leftmargin=1.2em]
  \item \textbf{\textbf{$\tau^2$-Bench} }
  We run 4 independent trials and report pass\^{}1 under temperature=0.0 (greedy decoding), with GPT-5.1~\citep{openai2025gpt51} serving as the user simulator.
  \item \textbf{\textbf{ACEBench} }
  Since the agent-task split contains only 50 instances, we repeat the evaluation 4 times and report the mean accuracy for stability.
  We use temperature=0.6 for inference, with GPT-4.1~\citep{openai2025gpt41} serving as the user simulator.
  \item \textbf{\textbf{BFCL-MT} }
  We run inference with temperature=0.6.
\end{itemize}

\paragraph{Non-agentic Evaluation.}
For both AIME2024 and AIME2025, we adopt a unified decoding protocol with temperature=0.6 and top-p=0.95.
To improve evaluation stability, we consider two top-k settings: k=20 and k=-1 (no top-k restriction).
For each benchmark, we perform 32 independent generations under each setting and estimate the pass rate by averaging correctness over samples.
The final reported score is the average of the two pass-rate estimates.

\subsubsection{Results}
\label{sec:eval_results}

\begin{table*}[t]
\centering
\caption{\textbf{Agentic Benchmark Results Across Training Stages.} Performance on BFCL-MT, $\tau^2$-Bench, and ACEBench for the Original, SFT, and RL models.}
\label{tab:stage_aligned_bfcl_tau2_acebench}
\footnotesize
\setlength{\tabcolsep}{2.2pt}
\renewcommand{\arraystretch}{1.15}
\resizebox{\textwidth}{!}{%
\begin{tabular}{@{}l lllll lll lll@{}}
\toprule
& \multicolumn{5}{c}{\textbf{BFCL-MT}}
& \multicolumn{3}{c}{\textbf{$\tau^2$-Bench}}
& \multicolumn{3}{c}{\textbf{ACEBench}} \\
\cmidrule(lr){2-6}\cmidrule(lr){7-9}\cmidrule(lr){10-12}
Model
& \textbf{Base} & \textbf{\shortstack{Missing\\Func}} & \textbf{\shortstack{Missing\\Param}} & \textbf{\shortstack{Long\\Context}} & \textbf{Overall}
& \textbf{ Retail} & \textbf{Telecom} & \textbf{Overall}
& \textbf{\shortstack{Multi\\Turn}} & \textbf{\shortstack{Multi\\Step}} & \textbf{Overall} \\
\midrule

\multicolumn{12}{@{}l}{\textbf{14B}} \\
Qwen3-14B
& 54.00 & 39.50 & 39.00 & 45.50 & \textbf{44.50}
& 55.00 & 34.10 & \textbf{44.55}
& 45.83 & 57.50 & \textbf{51.67} \\

Ours-14B$_{\mathrm{SFT}}$
& 67.50 & 25.50 & 48.00 & 53.00 &
\textbf{48.50}\textbf{\textcolor{green!50!black}{\scriptsize +4.00}}
& 63.80 & 37.10 &
\textbf{50.45}\textbf{\textcolor{green!50!black}{\scriptsize +5.90}}
& 51.67 & 83.75 &
\textbf{67.71}\textbf{\textcolor{green!50!black}{\scriptsize +16.04}} \\

Ours-14B$_{\mathrm{RL}}$
& 67.00 & 56.00 & 48.50 & 61.00 &
\textbf{58.13}\textbf{\textcolor{green!50!black}{\scriptsize +13.63}}
& 68.00 & 47.40 &
\textbf{57.70}\textbf{\textcolor{green!50!black}{\scriptsize +13.15}}
& 54.17 & 83.75 &
\textbf{68.96}\textbf{\textcolor{green!50!black}{\scriptsize +17.29}} \\
\midrule

\multicolumn{12}{@{}l}{\textbf{32B}} \\
Qwen3-32B
& 56.00 & 52.50 & 40.00 & 43.00 & \textbf{47.88}
& 64.70 & 34.70 & \textbf{49.70}
& 55.83 & 63.75 & \textbf{59.79} \\

Ours-32B$_{\mathrm{SFT}}$
& 67.00 & 40.00 & 46.00 & 55.50 &
\textbf{52.13}\textbf{\textcolor{green!50!black}{\scriptsize +4.25}}
& 66.70 & 37.90 &
\textbf{52.30}\textbf{\textcolor{green!50!black}{\scriptsize +2.60}}
& 56.67 & 78.75 &
\textbf{67.71}\textbf{\textcolor{green!50!black}{\scriptsize +7.92}} \\

Ours-32B$_{\mathrm{RL}}$
& 76.50 & 65.50 & 48.50 & 66.50 &
\textbf{64.25}\textbf{\textcolor{green!50!black}{\scriptsize +16.38}}
& 75.20 & 52.20 &
\textbf{63.70}\textbf{\textcolor{green!50!black}{\scriptsize +14.00}}
& 60.00 & 83.75 &
\textbf{71.88}\textbf{\textcolor{green!50!black}{\scriptsize +12.09}} \\

\bottomrule
\end{tabular}%
}
\end{table*}

As shown in Table~\ref{tab:bfcl_tau2_acebench_results}, we report performance on BFCL-MT, $\tau^2$-Bench, and ACEBench across multiple model scales, including closed-source models, open-source models, and our models. Our models achieve state-of-the-art results at matched parameter scales and are competitive with higher-parameter open-source and closed-source models on multiple metrics. Table~\ref{tab:stage_aligned_bfcl_tau2_acebench} further indicates that both the SFT and RL stages deliver consistent gains, with the RL stage contributing the largest improvement.

In addition, as shown in Table~\ref{tab:aime_decoding_robustness}, we evaluate AIME2024 and AIME2025 under two decoding settings for both 14B and 32B models. Although our method primarily optimizes agentic tool-use, it shows negligible degradation on non-agentic complex reasoning, indicating robust behavior.

\begin{table}[t]
\centering
\caption{\textbf{Non-agentic benchmark results}. We report AIME2024 and AIME2025 under two decoding settings for both 14B and 32B models. }
\label{tab:aime_decoding_robustness}
\footnotesize
\setlength{\tabcolsep}{4.2pt}
\renewcommand{\arraystretch}{1.15}
\begin{tabular}{@{}l ccc ccc@{}}
\toprule
& \multicolumn{3}{c}{\textbf{(topp=0.95, topk=20, temperature=0.6)}}
& \multicolumn{3}{c}{\textbf{(topp=0.95, topk=-1, temperature=0.6)}} \\
\cmidrule(lr){2-4}\cmidrule(lr){5-7}
\textbf{Model} & \textbf{AIME2024} & \textbf{AIME2025} & \textbf{Avg}
              & \textbf{AIME2024} & \textbf{AIME2025} & \textbf{Avg} \\
\midrule

Qwen3-14B
& 80.00 & 66.90 & 73.45
& 78.50 & 66.70 & 72.60 \\
ASTRA-14B-Thinking-v1
& 80.10 & 66.70 & 73.40
& 78.80 & 66.40 & 72.60 \\
\midrule

Qwen3-32B
& 83.00 & 66.80 & 74.90
& 82.40 & 65.90 & 74.15 \\
ASTRA-32B-Thinking-v1
& 81.40 & 68.30 & 74.85
& 81.20 & 69.10 & 75.15 \\

\bottomrule
\end{tabular}
\end{table}

\section{Discussion}
In this section, we conduct a comprehensive analysis to better understand the factors that shape effective tool-use behavior, focusing on three complementary perspectives: irrelevant-tool mixing strategies, reward design for RL training stability, and stage-wise analysis of agent behavior and performance across the original, SFT, and RL models.

\subsection{Improving Tool-Use Discrimination via Irrelevant Tool Mixing}

As described in Section~\ref{sec:irrelevant_tool_mixing}, ASTRA shapes tool-use behavior by exposing the agent to irrelevant tools during reinforcement learning. This design encourages the policy to learn not only correct tool usage but also effective tool discrimination.

\begin{figure}[h!]
    \centering
    \includegraphics[width=0.9\linewidth]{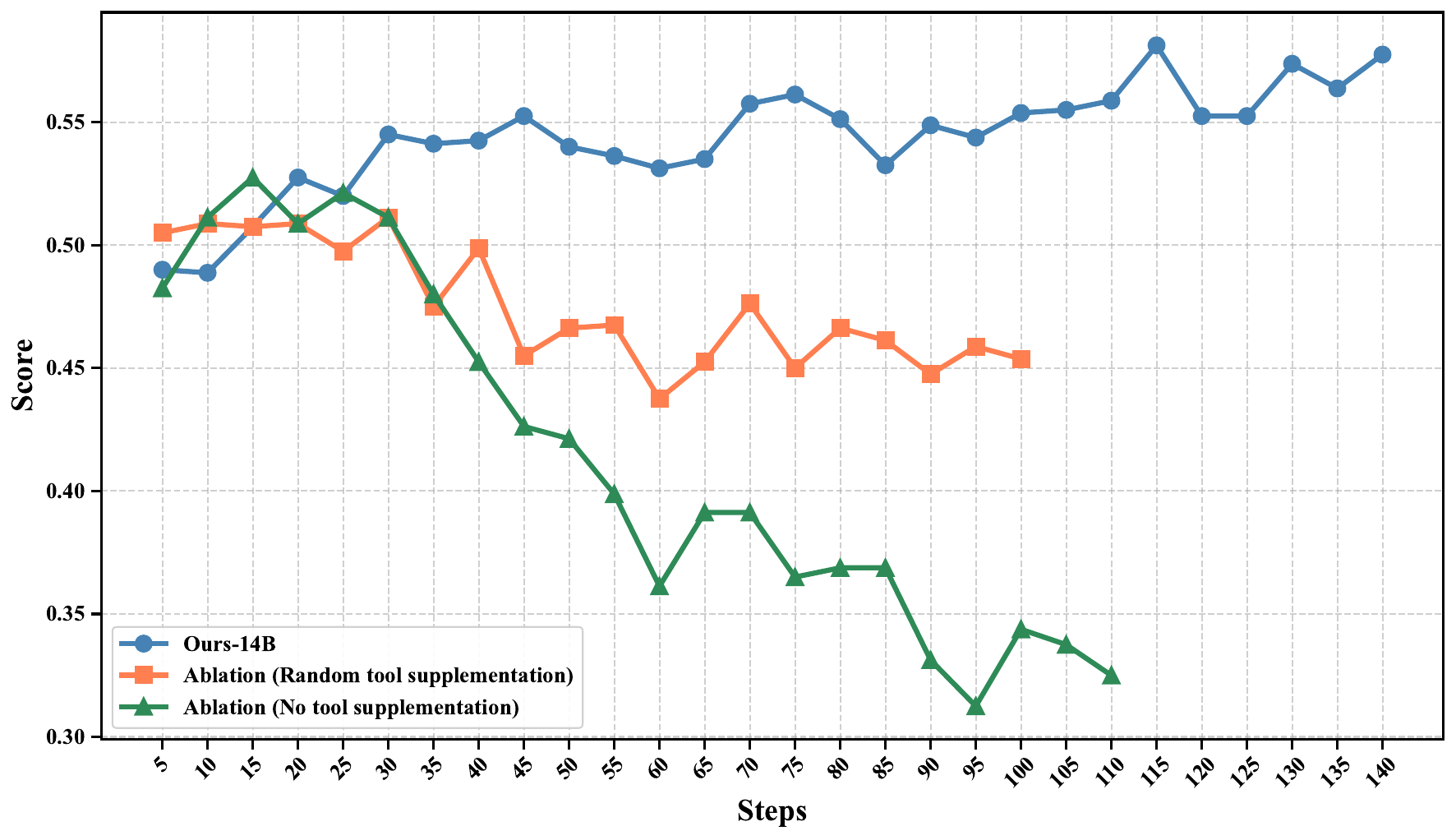}
    \caption{Ablation on Irrelevant-Tool Mixing Settings.}
    \label{fig:tool_ablation}
\end{figure}

To study how this interacts with reward optimization, we perform two ablations under identical RL settings, varying only the tool set composition: (i) \textbf{No Irrelevant Tools}, where only ground-truth tools are provided, and (ii) \textbf{Random Irrelevant Tools}, where 5--9 tools are randomly sampled from other domains.

Results are shown in Figure~\ref{fig:tool_ablation}. Removing irrelevant tools yields the worst performance, as the policy overfits to a narrow tool-selection pattern and lacks pressure to optimize the precision component of the reward. Randomly mixing irrelevant tools improves performance by introducing basic discrimination signals, but remains inferior to the full ASTRA setup.

These ablations highlight that irrelevant-tool mixing is a necessary signal for learning \textbf{negative tool judgment}. When no irrelevant tools are provided, the policy is never required to reject a plausible-but-wrong option, and therefore fails to acquire the capability to identify tools as irrelevant under realistic toolset exposure.

\begin{figure}[h!]
    \centering
    \includegraphics[width=0.9\linewidth]{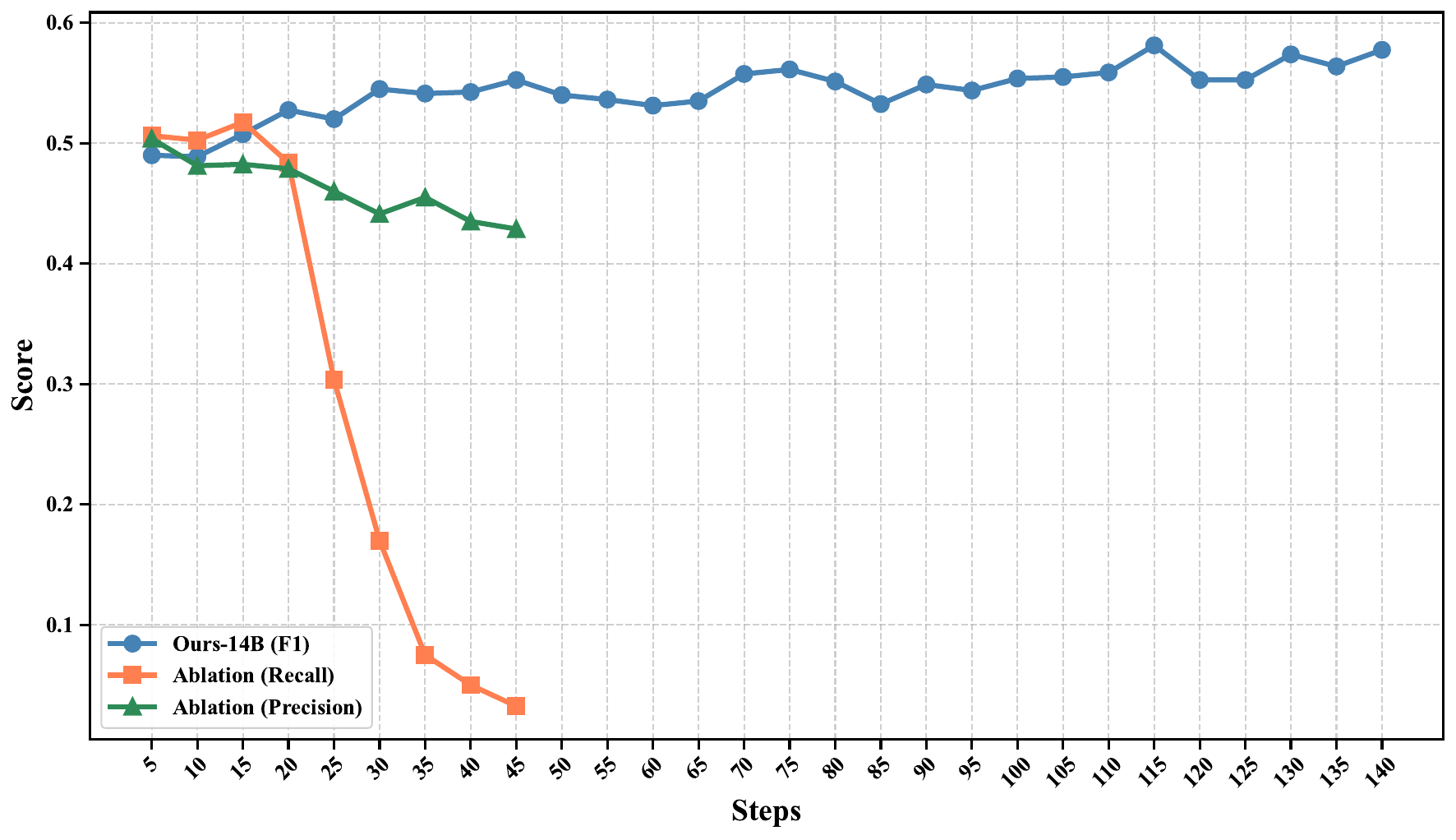}
    \caption{Ablation on Reward Configurations.}
    \label{fig:reward_ablation}
\end{figure}

Moreover, the effectiveness of irrelevant-tool mixing depends on the \textbf{similarity structure} among tools. In practice, a tool's similarity to other tools is not uniformly distributed across the dataset; some irrelevant tools are near-miss distractors while others are trivially dissimilar. The results suggest that a more balanced coverage over tool-similarity ranges---i.e., mixing irrelevant tools that span multiple similarity bands rather than concentrating on a single region---can better support reward optimization: it provides consistent discrimination pressure and enables the policy to learn both \textbf{which tools to call} and \textbf{which tools to ignore}.

\subsection{How Reward Design Shapes Tool-Use Behavior}
\label{sec:reward_design}

As described earlier, we use an F1-style trajectory reward to jointly encourage \textbf{task completion} and \textbf{interaction efficiency}. For a job with $n$ required sub-tasks, if the agent solves $\hat{n}$ sub-tasks with $c$ tool invocations, we define recall $r=\frac{\hat{n}}{n}$ and precision $p=\frac{\hat{n}}{c+\epsilon}$, and compute the reward as $\mathrm{F1}=\frac{2pr}{p+r}$.

\begin{figure}[h!]
    \centering
    \includegraphics[width=0.9\linewidth]{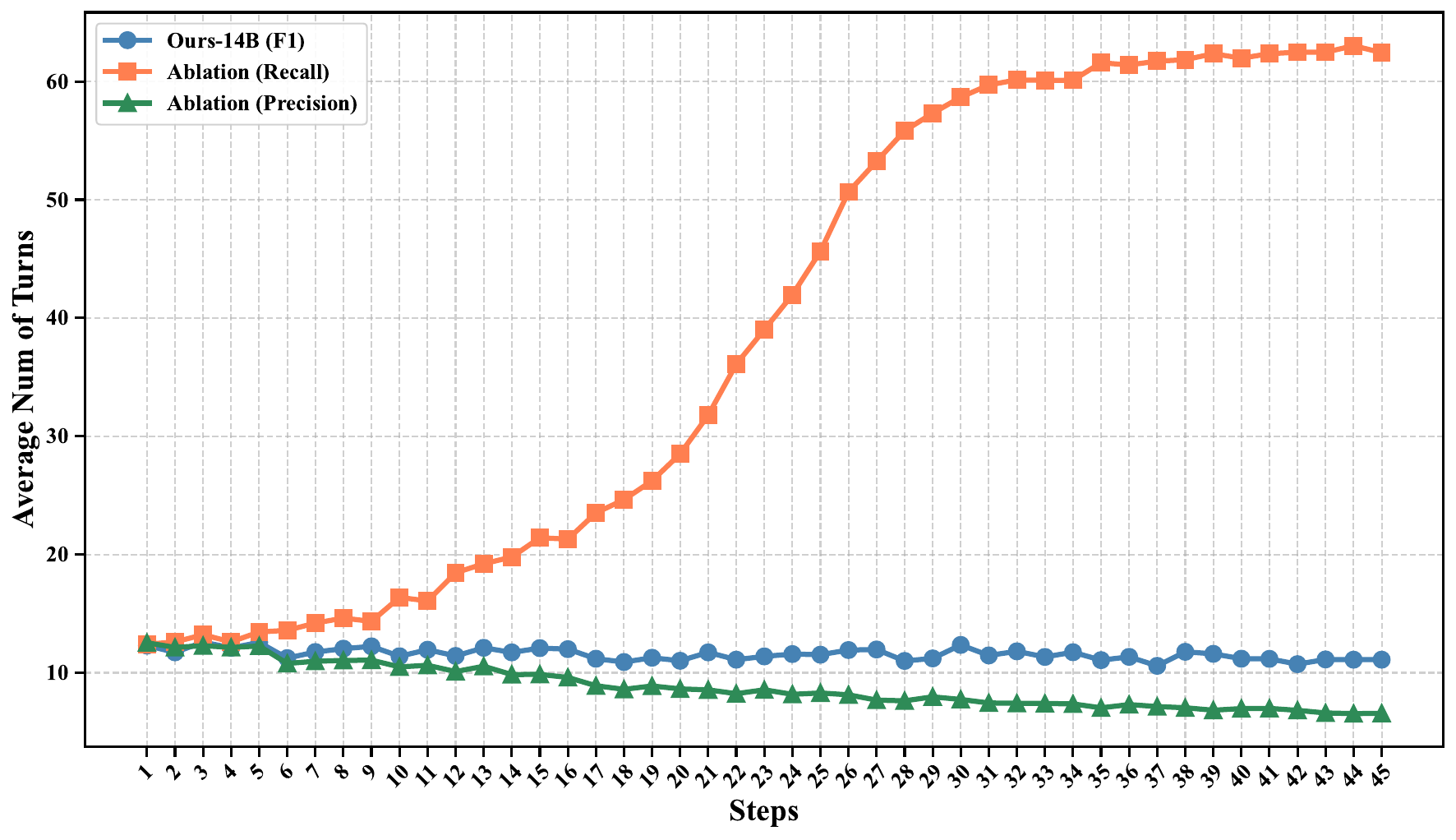}
    \caption{Dialogue-Turn Comparison Under Different Reward Configurations.}
    \label{fig:reward_ablation_turn}
\end{figure}

To study how reward shaping affects tool-use behavior, we run two ablations under the same RL configuration (identical initialization, data mixture, rollout settings, and GRPO optimization), changing only the trajectory reward: \textbf{recall-only} ($r$) and \textbf{precision-only} ($p$). Results are shown in Figure~\ref{fig:reward_ablation}.

We further examine training dynamics by tracking the number of interaction turns per trajectory across RL updates. As shown in Figure~\ref{fig:reward_ablation_turn}, the reward design induces markedly different turn-length distributions over training.

Optimizing recall-only quickly causes turns to explode, as the policy prolongs interaction and issues increasingly many tool calls, inflating sequence lengths and destabilizing online optimization until training collapses. In contrast, optimizing precision-only drives turns to drop sharply by discouraging tool calls. This pushes the policy toward overly conservative, short-horizon behavior that is brittle in multi-step settings and also collapses later in training.

Finally, the \textbf{F1} reward yields well-behaved turn distributions and stable training. By jointly optimizing recall and precision, it preserves an incentive for exploration (solving more sub-tasks via tool use) while simultaneously penalizing incorrect or irrelevant tool invocations through the precision term. This coupled objective provides a more balanced trade-off between \textbf{exploration} and \textbf{exploitation}, preventing both runaway tool overuse and overly conservative under-calling, and thereby supporting robust multi-step performance.

\subsection{Analysis at Different Training Stages}

\newcolumntype{Y}{>{\centering\arraybackslash}X}
\begin{table}[htbp]
\caption{Average Steps and Token Usage per Subtask on BFCL v3 MT.}
\label{tab:length_info}
\centering
\begin{tabularx}{\textwidth}{lYYY}
\toprule
Models & \makecell{Average steps \\ (per sub-job)} & \makecell{Average Tokens \\ (per step)} & \makecell{Average Tokens \\ (per sub-job)}  \\
\midrule
Qwen3-14B & 2.5 & 379.6 & 1096.7 \\
$\text{ASTRA-14B-Thinking-v1}_\text{SFT}$ & 3.1 & 171.6 & 686.4  \\
$\text{ASTRA-14B-Thinking-v1}_\text{RL}$ & 3.2 & 237.9 & 898.6  \\
\midrule
Qwen3-32B & 3.7 & 361.7 & 1145.3 \\
$\text{ASTRA-32B-Thinking-v1}_\text{SFT}$ & 3.1 & 192.0 & 672.1  \\
$\text{ASTRA-32B-Thinking-v1}_\text{RL}$ & 3.1 & 317.8 & 1130.1 \\
\bottomrule
\end{tabularx}
\end{table}

We compare behaviors and performance across three stages: the \textbf{original model}, the model after \textbf{SFT}, and the final model after \textbf{RL}. 

\paragraph{Behaviors Analysis.}
We analyze two high-level statistics: interaction steps and average output length. The results are shown in Table~\ref{tab:length_info}.

\begin{itemize}[leftmargin=1.2em]
\item \textbf{Interaction Steps } The average number of interaction steps remains largely unchanged across all stages. Neither SFT nor RL systematically alters dialogue depth, indicating that performance differences are not driven by trivial changes in interaction length.

\item \textbf{Output Length } Output length shows a clear stage-wise pattern. The original model generates the \textbf{longest trajectories}, while SFT produces the \textbf{shortest outputs} by compressing reasoning into concise, demonstration-style patterns. The RL-trained model converges to an \textbf{intermediate length}, longer than SFT but shorter than the original model.
\end{itemize}

\paragraph{Performance Analysis.}
Table~\ref{tab:stage_aligned_bfcl_tau2_acebench} summarizes agentic benchmark performance across training stages.
Overall, both \textbf{SFT} and \textbf{RL} improve over the \textbf{original model}, with \textbf{RL} consistently achieving the best results.

\begin{itemize}[leftmargin=1.2em]
\item \textbf{SFT improves multi-turn tool-use adaptation }
By using \textbf{tool-chain-based pipeline}, SFT provides a strong cold start by teaching structured tool invocation, multi-turn state tracking, and adherence to interaction conventions. This initialization consistently outperforms the original model. 

\item \textbf{RL further improves performance via broader exploration }
Compared to SFT, RL delivers substantial additional gains.
We attribute this improvement to our \textbf{QA-based} RL method--instead of optimizing toward a single golden tool-sequence answer, we anchor supervision on sub-QA pairs. This method encourages the model to search over a larger space of feasible trajectories that achieve the same intermediate subgoals and final-answer constraints. Consequently, the policy is optimized over a \textbf{more expansive search space with semantic and topological structure} for multi-turn tool usage, enabling trajectory-level credit assignment and recovery from suboptimal decisions in a more constrained topological search space.

\end{itemize}

\section{Related Work}

\subsection{Tool-Use Trajectory Synthesis}

Recent progress in tool-augmented language model agent has driven growing interest in systematically constructing large-scale, high-quality trajectories. 
A major line of work~\citep{qin2023toolllmfacilitatinglargelanguage,liu2025toolacewinningpointsllm} constructs large tool-centric corpora over extensive tool inventories to improve data diversity and scale. Subsequent efforts extend this paradigm to multi-turn settings by explicitly modeling tool-call sequences with executability constraints and verification mechanisms~\citep{yin2025magnetmultiturntoolusedata,zeng2025toolacemtnonautoregressivegenerationagentic}
Notably, APIGen-MT~\citep{prabhakar2025apigenmtagenticpipelinemultiturn} formalizes multi-turn trajectory synthesis through a two-phase framework that decouples task blueprint generation from simulated human--agent interactions, enabling controllable trajectory construction and demonstrating strong performance on agentic benchmarks.
Complementary approaches further broaden coverage by harvesting and standardizing tools from large-scale tool ecosystems or real-world tool servers~\citep{xu2025toucansynthesizing15mtoolagentic}.   

Beyond explicit tool inventories, another emerging line of work reduces reliance on predefined tools by converting implicit procedural knowledge in open-domain text into trainable multi-turn tool-use trajectories. GEM~\citep{xu2026unlockingimplicitexperiencesynthesizing} adopts a Text-to-Trajectory paradigm that extracts latent workflows from raw text and grounds them into executable trajectories, offering a scalable data source and improved cross-domain generalization. LoopTool~\citep{zhang2025looptoolclosingdatatrainingloop} closes the data--training loop by iteratively adapting the data distribution to model weaknesses, thereby improving robustness and long-horizon tool-use performance to overcome the limitations of one-shot offline synthesis.

\subsection{Environment Scaling}
As training and evaluating increasingly capable agents require exposure to diverse, executable, and stateful environments, a growing body of work focuses on scaling tool-interactive environments. However, most existing approaches rely on manual environment design, such as interactive benchmarks and controlled task suites~\citep{yao2024taubenchbenchmarktoolagentuserinteraction,barres2025tau2benchevaluatingconversationalagents,he2025vitabenchbenchmarkingllmagents}, which inherently constrains domain diversity and scalability due to the high cost of human design and maintenance.

To address these limitations, recent work has shifted toward programmatic environment construction to support scalable training.
EnvScaler~\citep{song2026envscalerscalingtoolinteractiveenvironments} addresses this challenge by automatically synthesizing tool-interactive environments, constructing executable environment skeletons and generating diverse scenarios with rule-based validators, thereby substantially expanding environment scale while preserving verifiability for both SFT and RL. AutoForge~\citep{cai2025autoforgeautomatedenvironmentsynthesis} further improves efficiency by synthesizing environments directly from tool documentation and introducing environment-level optimization to mitigate noisy simulated users.

Beyond direct environment synthesis, AgentScaler~\citep{fang2025generalagenticintelligenceenvironment} advances an alternative abstraction by modeling environments as read--write databases, enabling the generation of verifiable agent experiences under a unified interface together with a two-stage learning strategy. Complementary to environment construction, CuES~\citep{mai2025cuescuriositydrivenenvironmentgroundedsynthesis} tackles task scarcity through curiosity-driven, environment-grounded task synthesis, extracting executable tasks from exploration trajectories without predefined goals. Finally, GenEnv~\citep{guo2025genenvdifficultyalignedcoevolutionllm} frames training as a co-evolutionary process, where the environment generator serves as a curriculum policy that dynamically aligns task difficulty with the agent’s evolving capability region.


\section{Conclusion and Future Work}

In this work, we presented ASTRA, a fully automated, end-to-end framework for training tool-augmented language model agents via scalable data synthesis and rule-verifiable multi-turn reinforcement learning. ASTRA unifies multi-turn trajectory synthesis that leverages the static topology of tool-call graphs for supervised fine-tuning with environment synthesis that captures the rich, compositional topology of human semantic reasoning, producing independent, executable, and rule-verifiable Python environments for online RL. 

Across multiple agentic tool-use benchmarks, ASTRA-trained models achieve strong performance at comparable scales while preserving general reasoning ability. We open-source the data synthesis pipelines, synthesized environments, and trained models to support reproducibility and future research. 
We anticipate that ASTRA can help alleviate practical deployment bottlenecks by reducing reliance on static, scenario-specific labeled data. Instead, it may be possible to synthesize multiple executable environments per scenario and train agents via iterative interaction to improve robustness in downstream applications.

Multi-turn, user-interactive agents are increasingly important in real-world deployments. In future work, we will extend ASTRA to incorporate multi-turn user interaction during training and evaluation, improving robustness to evolving intents and interactive feedback while maintaining verifiability and reproducibility. More broadly, scalable deployment also requires cost-aware automation. Since executable environment synthesis can be expensive, we will explore refining and verifying the QA-derived topology prior to code generation, using the validated topology as prior information and instantiating code environments only for high-confidence specifications.

\clearpage
\section{Contribution}
\label{sec:contribution}
\noindent
\begin{minipage}[t]{0.48\textwidth}
  \subsection*{Core Contributors}\vspace{3pt}
  \begin{itemize}[label={}, leftmargin=0pt, itemsep=0pt, topsep=2pt]
  
    \item Xiaoyu Tian
    \item Haotian Wang
    \item Shuaiting Chen
    \item Hao Zhou
    \item Kaichi Yu
    \item Yudian Zhang
    \item Jade Ouyang
    \item Junxi Yin
    \item Jiong Chen
    
  \end{itemize}
\end{minipage}
\hfill
\begin{minipage}[t]{0.48\textwidth}
  \subsection*{Contributors}\vspace{3pt}
  \begin{itemize}[label={}, leftmargin=0pt, itemsep=0pt, topsep=2pt]

    \item Baoyan Guo
    \item Lei Zhang
    \item Junjie Tao
    \item Yuansheng Song
    \item Ming Cui
    \item Chengwei Liu
    
  \end{itemize}
\end{minipage}

\clearpage

\newpage
\bibliographystyle{unsrt}
\bibliography{reference}

\newpage
\appendix

\section{Appendix}

\definecolor{userbg}{RGB}{240, 248, 255}   
\definecolor{badbg}{RGB}{255, 235, 235}    
\definecolor{goodbg}{RGB}{235, 255, 235}   
\definecolor{bordergray}{RGB}{180, 180, 180} 

\definecolor{toolcolor}{RGB}{0,102,204} 
\newcommand{\Tool}[1]{%
  {\textcolor{toolcolor}{\texttt{\bfseries #1}}}%
}

\newlength{\CaseBlockSep}
\setlength{\CaseBlockSep}{2pt} 

\newcommand{\UserBox}[1]{%
    \vspace{\CaseBlockSep} 
    {\noindent\bfseries\sffamily User Question} 
    \par \vspace{0.1cm} 
    \begin{tcolorbox}[
        enhanced,
        colback=userbg,
        colframe=bordergray,
        boxrule=0.5pt,
        arc=2mm,
        top=2mm, bottom=2mm, left=2mm, right=2mm,
        width=\linewidth,
        nobeforeafter,
        fontupper=\sffamily
    ]
    #1
    \end{tcolorbox}
}

\newcommand{\BadModelBox}[2]{
    \vspace{\CaseBlockSep} 
    {\noindent\bfseries\sffamily #1 \hspace{0.5em} \textcolor{red}{\Large \ding{55}}} 
    \par \vspace{0.1cm} 
    \begin{tcolorbox}[
        enhanced,
        colback=badbg,
        colframe=bordergray,
        boxrule=0.5pt,
        arc=2mm,
        top=2mm, bottom=2mm, left=2mm, right=2mm,
        width=\linewidth,
        nobeforeafter,
        fontupper=\sffamily
    ]
    #2
    \end{tcolorbox}
}

\newcommand{\GoodModelBox}[2]{
    \vspace{\CaseBlockSep}
    {\noindent\bfseries\sffamily #1 \hspace{0.5em} \textcolor{green!60!black}{\Large \ding{51}}}
    \par \vspace{0.1cm}
    \begin{tcolorbox}[
        enhanced,
        colback=goodbg,
        colframe=bordergray,
        boxrule=0.5pt,
        arc=2mm,
        top=2mm, bottom=2mm, left=2mm, right=2mm,
        width=\linewidth,
        nobeforeafter,
        fontupper=\sffamily
    ]
    #2
    \end{tcolorbox}
}

\subsection{Data Analysis}
\label{sec:data_analysis}

\subsubsection{SFT Data}

\begin{figure}[htbp]
    \centering
    \includegraphics[width=1.0\textwidth]{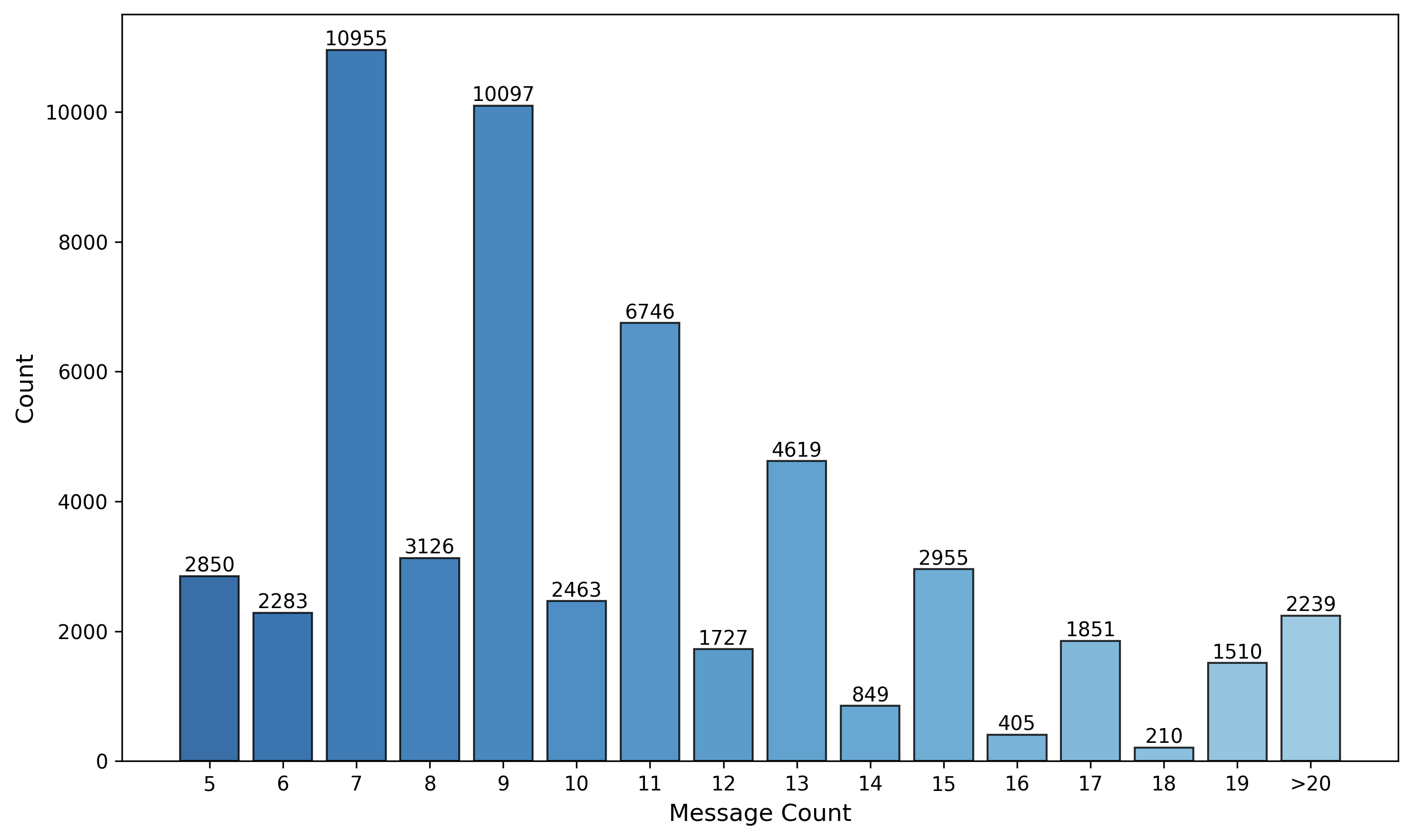}
    \caption{Distribution of Messages per Sample in SFT.}
    \label{fig:sft_01_messages_per_sample}
\end{figure}

We construct a high-quality SFT dataset comprising 54,885 multi-turn conversation samples with a total of 580,983 messages. Each sample contains an average of 10.59 messages, with the detailed distribution shown in Figure~\ref{fig:sft_01_messages_per_sample}.

\begin{figure}[htbp]
    \centering
    \includegraphics[width=1.0\textwidth]{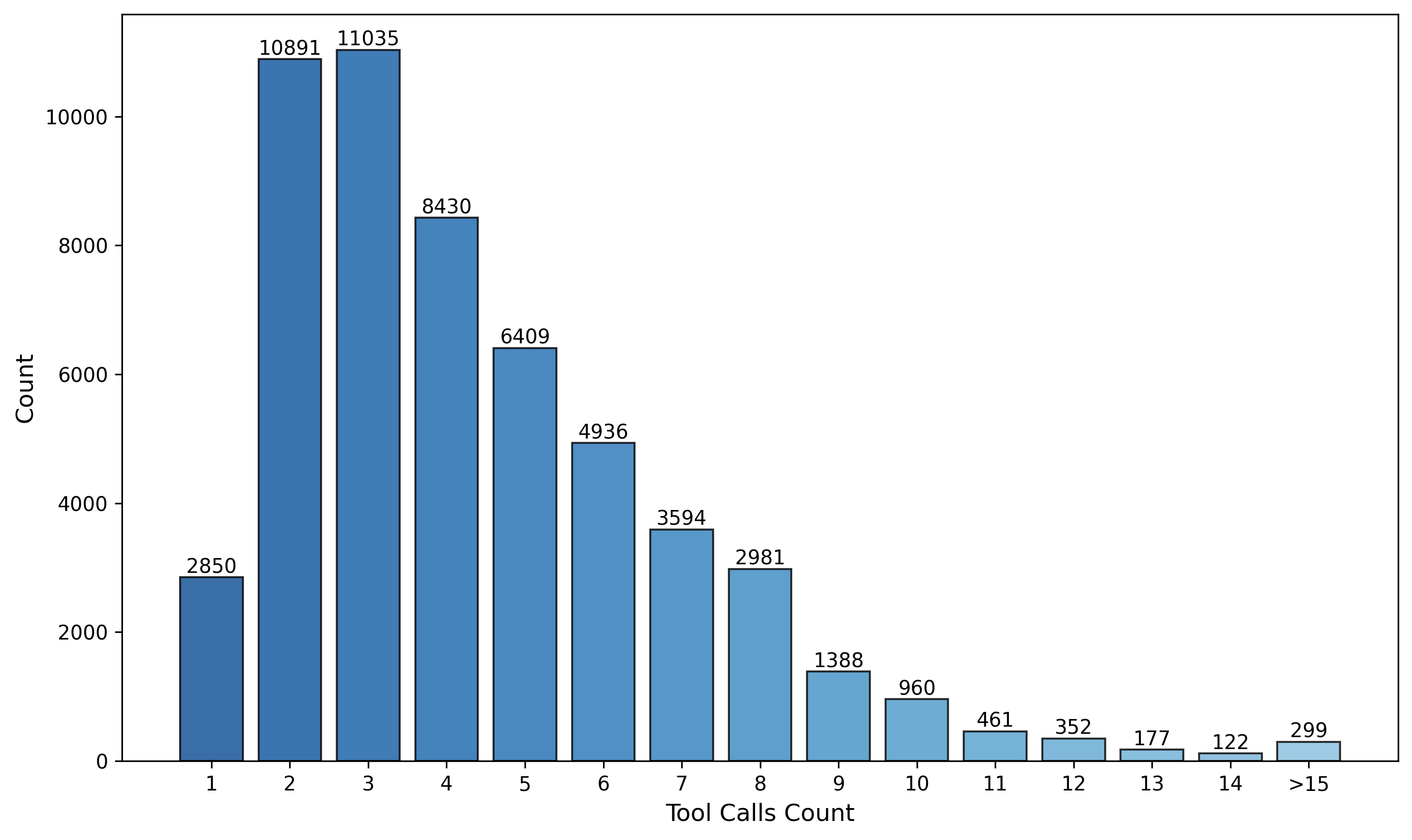}
    \caption{Distribution of the Number of Tool Calls per sample in SFT.}
    \label{fig:sft_tool_calls_distribution}
\end{figure}

All samples involve tool calling, with an average of 4.42 tool invocations per conversation; 72.2\% of the samples contain 1–5 tool calls, as illustrated in Figure~\ref{fig:sft_tool_calls_distribution}. The role distribution is dominated by tool responses (41.8\%) and assistant utterances (39.3\%), reflecting the tool-intensive nature of the dataset. These tool calls span 6,765 unique tool functions, covering reasoning, computation, search capabilities, and more.

\subsubsection{RL Data}
\begin{figure}[htbp]
\centering
\includegraphics[width=1.0\linewidth]{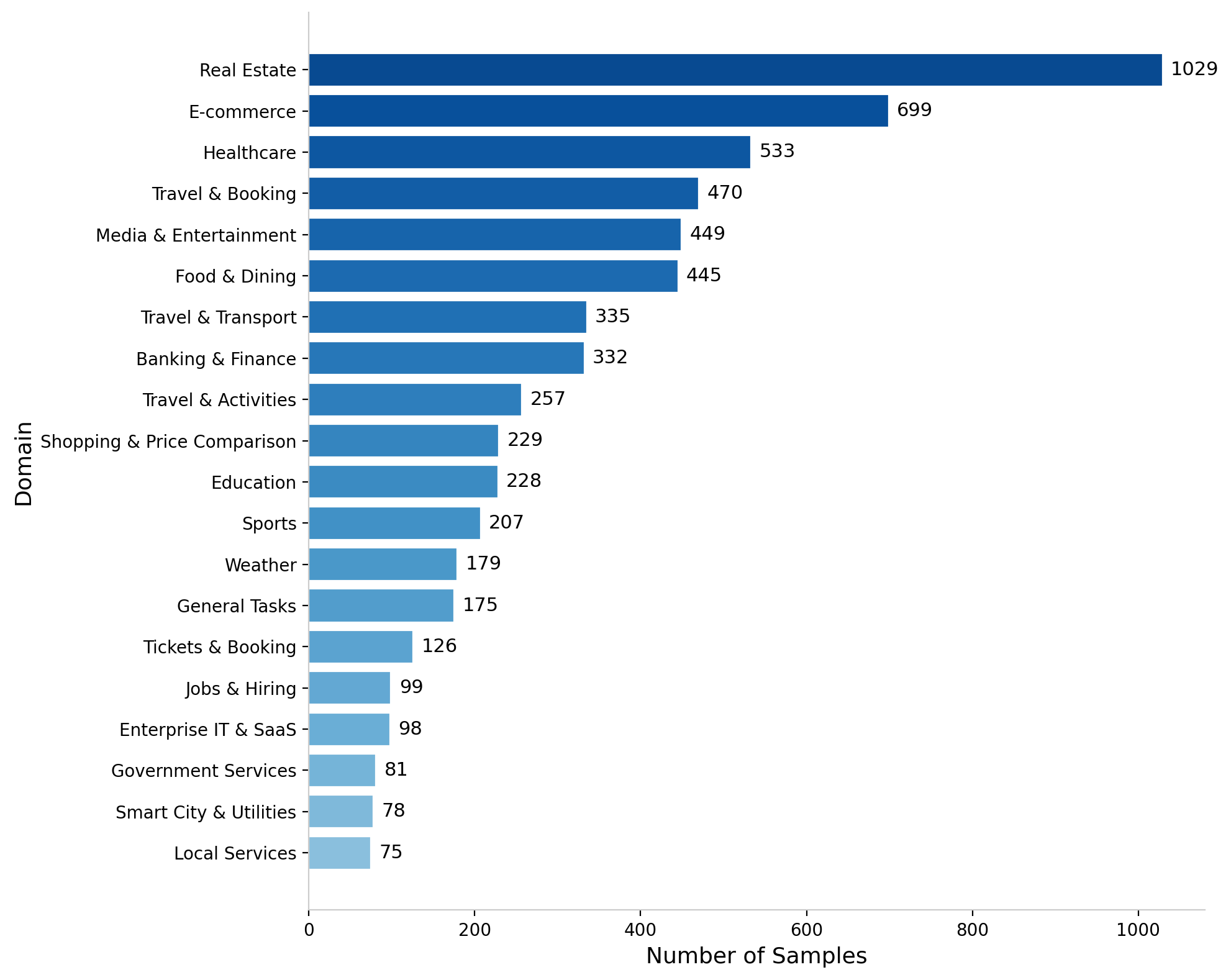}
\caption{Top 20 Domain Distribution in the RL Dataset.}
\label{fig:rl_domain_distribution}
\end{figure}

Our RL dataset comprises 6,596 samples spanning diverse domains. As illustrated in Figure~\ref{fig:rl_domain_distribution}, the distribution is led by Real Estate (15.6\%), E-commerce (10.6\%), Healthcare (8.1\%), covering varied application scenarios. The collection is bilingual, with English samples accounting for 71.2\% (4,694) and Chinese samples for 28.8\% (1,902). 

\begin{figure}[htbp]
\centering
\includegraphics[width=1.0\linewidth]{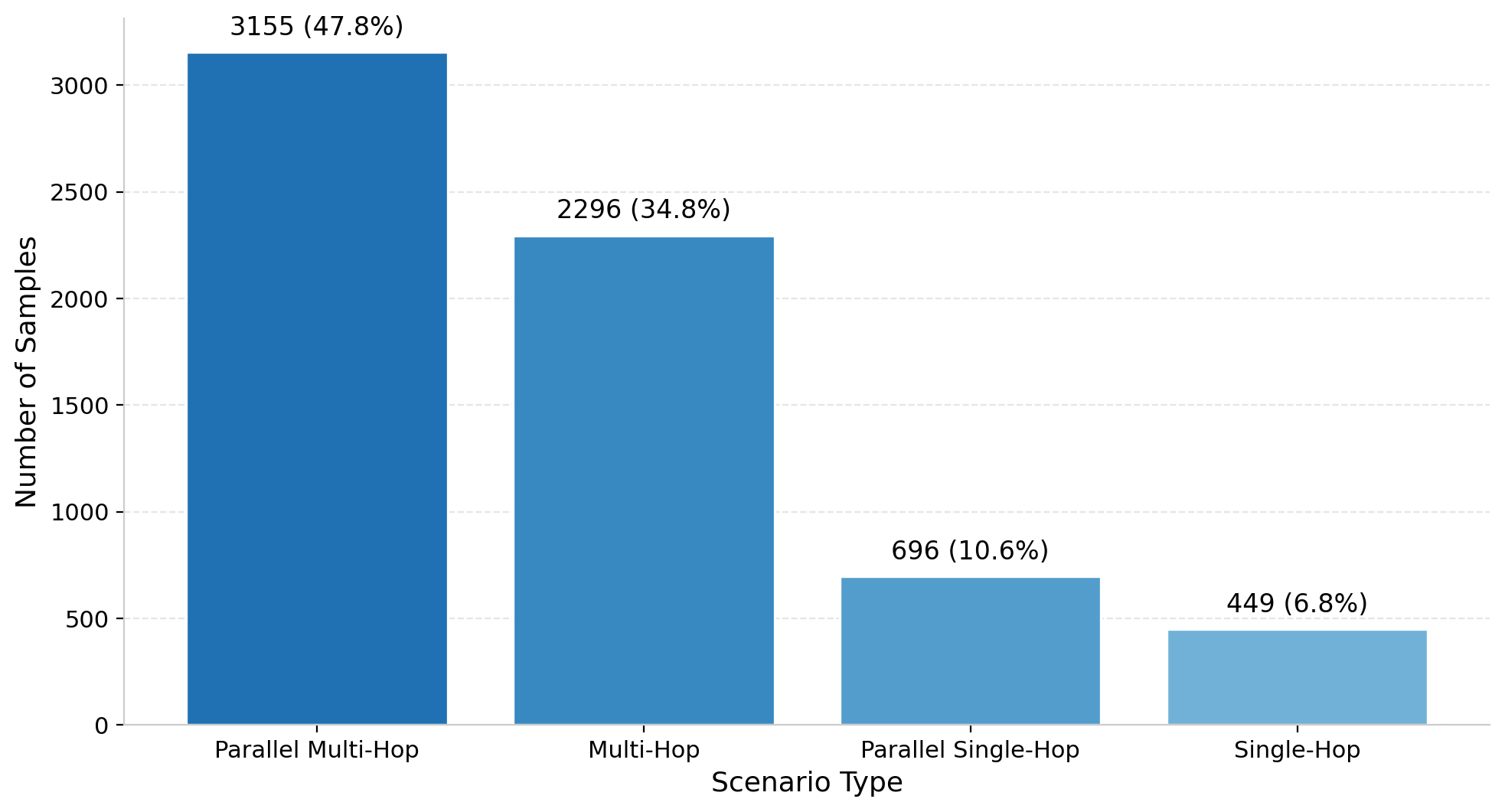}
\caption{Distribution of Scenario Types in RL.}
\label{fig:rl_scenario_type_distribution}
\end{figure}

\begin{figure}[htbp]
\centering
\includegraphics[width=1.0\linewidth]{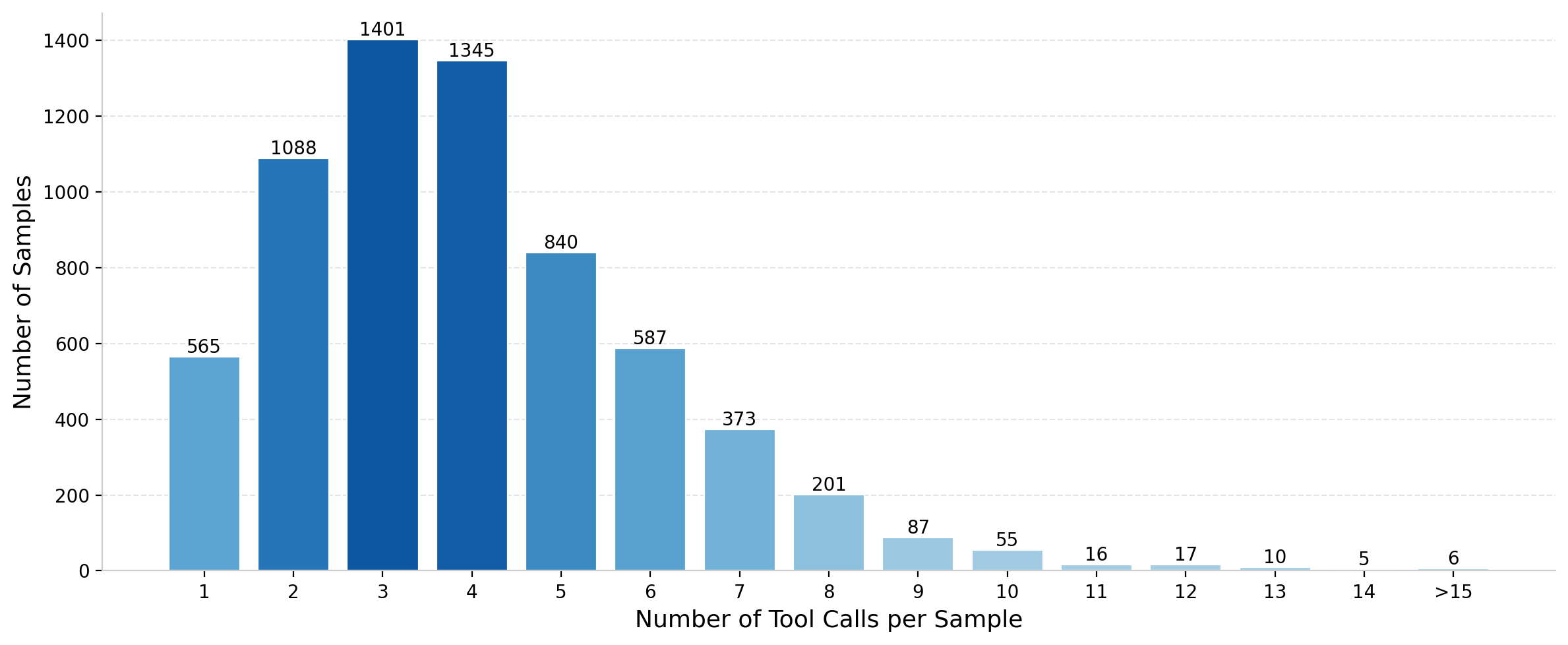}
\caption{Distribution of Tool Calls per sample in RL.}
\label{fig:rl_tools_per_sample_distribution}
\end{figure}

In terms of task complexity, samples contain an average of 4.37 reasoning hops (median: 4.0; range: 1--20). The distribution of scenario types, shown in Figure~\ref{fig:rl_scenario_type_distribution}, reveals that Parallel Multi-Hop scenarios are the most prevalent (47.8\%), followed by Multi-Hop (34.8\%), Parallel Single-Hop (10.6\%), and Single-Hop (6.8\%), underscoring a focus on complex, multi-step reasoning. At the sub-question level, 91.3\% of the 28,794 total sub-questions require external tool calls. As detailed in Figure~\ref{fig:rl_tools_per_sample_distribution}, most samples necessitate 2--5 tool calls, with an average of 3.98 calls per sample. 

Furthermore, the reasoning structure indicates that 44.2\% of steps can be parallelized, while 55.8\% require serial execution due to data dependencies. 

\clearpage

\subsection{Case study}
\label{sec:case_study}

\begin{tcolorbox}[
    title=The Original Model Is Incorrect; the SFT and RL Models Are Correct,
    breakable,
    enhanced,
    colback=white,
    colframe=gray!80!black,
    before skip=0pt,
    width=\dimexpr\linewidth+2cm\relax, 
    enlarge left by=-0.75cm,
    enlarge right by=-0.75cm,
    left=2mm, right=2mm, top=2mm, bottom=2mm,
    before upper=\linespread{1.05}\selectfont
]

    \UserBox{
        I'm interested in keeping up with Lulu Qian and her impressive achievements in the tech world, even if it's all a bit over my head! One company I've heard about lately that caught my attention because of its innovative tech solutions is ``Quasar Ltd.''. I've become quite intrigued by what they offer. Could you help me out by finding the current price and trading details for ``Quasar Ltd.''. so I can better understand their market position?
    }

    \BadModelBox{Original Model Global Think}{
        Okay, let's see. The user wants information about ``Quasar Ltd.'', specifically the current price and trading details to understand their market position. First, I need to figure out the stock symbol for ``Quasar Ltd.''. The user mentioned the company name, but not the symbol. So, I should use the \Tool{get\_symbol\_by\_name} function to get the stock symbol.
        \vspace{4pt}
        
        Once I have the symbol, the next step is to retrieve the current stock information using \Tool{get\_stock\_info} with the symbol obtained. That should give details like price, percent change, volume, and moving averages. Also, the user might want to add this stock to their watchlist for tracking, so I should check if they want to use \Tool{add\_to\_watchlist} afterwards. But the user hasn't explicitly asked to add it yet, so maybe after providing the info, I can suggest adding it.
        \vspace{4pt}
        
        Wait, the user's main request is to find the current price and trading details. So first priority is getting the symbol, then the stock info. Let me make sure to call those two functions. If the symbol isn't found, I'll need to inform the user. But assuming it's a real company, proceed. Let's start with getting the symbol.
    }

    \GoodModelBox{SFT Model Global Think}{
        The user is asking for information about ``Quasar Ltd.'' stock. They want to know the current price and trading details to understand the company's market position.
        \vspace{4pt}
        
        To help with this, I need to:
        \begin{enumerate}[leftmargin=*, nosep]
            \item First get the stock symbol for ``Quasar Ltd.'' using the \Tool{get\_symbol\_by\_name} function
            \item Then get the stock details using the \Tool{get\_stock\_info} function with that symbol
        \end{enumerate}
        \vspace{4pt}
        
        Let me start by getting the symbol for ``Quasar Ltd.''.
    }

    \GoodModelBox{RL Model Global Think}{
        The user wants to find the current price and trading details for ``Quasar Ltd.''. 
        \vspace{4pt}
        
        Looking at the available functions, I need to:
        \begin{enumerate}[leftmargin=*, nosep]
            \item First, get the stock symbol for ``Quasar Ltd.'' using the \Tool{get\_symbol\_by\_name} function
            \item Then, get the stock details using the \Tool{get\_stock\_info} function with that symbol
        \end{enumerate}
        \vspace{4pt}
        
        Let me check the parameters:
        \begin{itemize}[leftmargin=*, nosep]
            \item \Tool{get\_symbol\_by\_name}: name (required) -- ``Quasar Ltd.''
            \item \Tool{get\_stock\_info}: symbol (required) -- the symbol returned from the first function call
        \end{itemize}
        \vspace{4pt}
        
        So I'll first call \Tool{get\_symbol\_by\_name} with ``Quasar Ltd.'' and then use that symbol to call \Tool{get\_stock\_info}.
    }

\end{tcolorbox}

\clearpage

\begin{tcolorbox}[
    title=The Original and SFT Models Is Incorrect; the RL Model Are Correct,
    breakable,
    enhanced,
    colback=white,
    colframe=gray!80!black,
    before skip=0pt,
    width=\dimexpr\linewidth+2cm\relax, 
    enlarge left by=-0.75cm,
    enlarge right by=-0.75cm,
    left=2mm, right=2mm, top=2mm, bottom=2mm,
    before upper=\linespread{1.05}\selectfont
]

    \UserBox{
        I'm gearing up for a quick business getaway and need my ride all set. Would you be able to verify if my tire pressure is in check? If it falls under 37.5 PSI, perhaps we could swing by the nearest tire shop?
    }

    \BadModelBox{Original Model Global Think}{
        Okay, the user is preparing for a business trip and wants to ensure their car is ready. They specifically mentioned checking tire pressure and visiting a tire shop if any tire is below 37.5 PSI.
        \vspace{4pt}
        
        First, I need to check the tire pressure using the \Tool{check\_tire\_pressure} function. This function returns the pressure for each tire and whether it's healthy. 
        \vspace{4pt}
        
        Once I get the data, I'll review each tire's PSI. If any are below 37.5, the healthy\_tire\_pressure field might indicate false. If that's the case, I should use \Tool{find\_nearest\_tire\_shop} to locate a shop for them. 
        \vspace{4pt}
        
        Wait, even if healthy is false, maybe the user wants to know the exact pressures. So I should list each tire's pressure. But the function's response includes healthy\_tire\_pressure as a boolean. If it's false, then proceed to find the shop.
        \vspace{4pt}
        
        So the steps are: call \Tool{check\_tire\_pressure}, evaluate the result, and if needed, call \Tool{find\_nearest\_tire\_shop}. Let's start with the first function.
    }

    \BadModelBox{SFT Model Global Think}{
        The user wants me to:
        \begin{enumerate}[leftmargin=*, nosep]
            \item Check the tire pressure of their vehicle
            \item If any tire pressure is under 37.5 PSI, find the nearest tire shop
        \end{enumerate}
        \vspace{4pt}
        
        Let me start by checking the tire pressure first, then based on the results, I'll decide whether to find a tire shop.
        \vspace{4pt}
        
        I'll use the \Tool{check\_tire\_pressure} function to get the current tire pressure readings.
    }

    \GoodModelBox{RL Model Global Think}{
        The user wants me to check their tire pressure and if it's under 37.5 PSI, find the nearest tire shop.
        \vspace{4pt}
        
        Let me break this down:
        \begin{enumerate}[leftmargin=*, nosep]
            \item First, I need to check the tire pressure using the \Tool{check\_tire\_pressure} function.
            \item Then, I need to evaluate the results. If any tire is under 37.5 PSI, I should use the \Tool{find\_nearest\_tire\_shop} function.
        \end{enumerate}
        \vspace{4pt}
        
        Let me start with checking the tire pressure.
        \vspace{4pt}
        
        Looking at the \Tool{check\_tire\_pressure} function:
        \begin{itemize}[leftmargin=*, nosep]
            \item It has no required parameters
            \item It returns a dict with tire pressure info including \texttt{\textbf{frontLeftTirePressure}}, \\ \texttt{\textbf{frontRightTirePressure}}, \texttt{\textbf{rearLeftTirePressure}}, \texttt{\textbf{rearRightTirePressure}}, \\ and \texttt{\textbf{healthyTirePpressure}}
        \end{itemize}
        \vspace{4pt}
        
        So I'll call that first, then based on the results, decide if I need to find a tire shop.
        \vspace{4pt}
        
        Let me make the function call for checking tire pressure.
    }

\end{tcolorbox}

\subsection{Example Prompts}
\label{sec:Key Prompt}
Here we list four prompts used in ASTRA.
\medskip

\begin{tcolorbox}[
    title={Example Prompt 1: Query Generating from Tool Chain},
    breakable,
    enhanced,
    colback=white,
    colframe=gray!80!black,
    before skip=0pt,
    width=\dimexpr\linewidth+2cm\relax, 
    enlarge left by=-0.75cm,
    enlarge right by=-0.75cm,
    left=2mm, right=2mm, top=2mm, bottom=2mm,
    before upper=\linespread{0.95}\selectfont
]
You are a helpful AI assistant.

\medskip
You have the following tools:

\begin{verbatim}
{TOOLS}
\end{verbatim}

These tools are designed for the following scenario:

\begin{verbatim}
{SCENERY}
\end{verbatim}

\medskip
I will provide you with a tool invocation chain (i.e., the sequence and content of tool calls). Your task is:

\begin{enumerate}
    \item Determine whether this tool invocation chain is meaningful and valid (i.e., whether it can accomplish a reasonable task).
    \item If it is meaningful, extract the original user intent or query corresponding to this chain.
    \item Output the result strictly in the specified JSON format:

\begin{verbatim}
Example output:
{"valid": true, "query": "Check the weather in Beijing."}
\end{verbatim}

\end{enumerate}

\medskip
Field description:

\begin{itemize}
    \item \texttt{valid}: Boolean value indicating whether the chain is valid (true = valid, false = invalid).
    \item \texttt{query}: String representing the user's instruction or intent; leave it as an empty string if invalid.
\end{itemize}

Please follow the above format strictly and do not include any extra explanations or text.

\medskip
Input:

\begin{verbatim}
<ToolInvocationChain>{chain}</ToolInvocationChain>
\end{verbatim}

\medskip
Output:

\end{tcolorbox}

\begin{tcolorbox}[
    title={Example Prompt 2: Tool Call Conciseness},
    breakable,
    enhanced,
    colback=white,
    colframe=gray!80!black,
    before skip=0pt,
    width=\dimexpr\linewidth+2cm\relax, 
    enlarge left by=-0.75cm,
    enlarge right by=-0.75cm,
    left=2mm, right=2mm, top=2mm, bottom=2mm,
    before upper=\linespread{0.95}\selectfont
]
\textbf{Task Overview}

You will act as an evaluator and read a complete trajectory of an Agent interacting with tools. Your goal is to assess the \textbf{cost efficiency} of each tool call, focusing on whether the call is concise, whether it could be optimized to reduce the number of calls or cost, and to identify redundant, repetitive, or inefficient calling paths.

\medskip
\textbf{Input Format}
\begin{enumerate}
    \item The trajectory includes: the user’s original request, the Agent’s reasoning, each tool call (with parameters), tool responses, and the Agent’s final reply.
    \item The trajectory is provided as a JSON object with the following structure:
\end{enumerate}

\begin{verbatim}
{
"tools": [...],
"messages": [
    {
    "role": "system" | "user" | "assistant" | "tool",
    "content": "...",
    "tool_calls": [...]
    }
]
}
\end{verbatim}
The complete "trajectory" will be provided at the end of the prompt via the "{trajectory}" placeholder.

\medskip
\textbf{Core Metric}

\medskip
Each tool call is scored as 0 or 1:

\begin{itemize}
    \item \textbf{1.0 (Necessary)}: Without this call, the user’s goal cannot be achieved or would be very difficult to achieve; the call follows the context, parameters match the requirement, and it obtains new information or advances the process at minimal cost.
    \item \textbf{0.0 (Redundant)}: The call is disconnected from the goal (including task substitution), repeatedly retrieves already available information, has incorrect parameters (including format errors), involves \textbf{parameter hallucination} (the model fabricates parameters not provided by the user or uses default/example parameters), or represents invalid retries.
\end{itemize}

\textbf{General Rules}
\begin{enumerate}
    \item \textbf{Necessity Principle}: If the call is unrelated or the goal is already achieved, score 0.
    \item \textbf{Information Gain Principle}: If the call does not provide critical missing info, score 0.
    \item \textbf{Reasonable Orchestration Principle}:
    \begin{itemize}
        \item \textbf{Batch Support}: Multiple serial calls when batch is supported → redundant.
        \item \textbf{Avoid Fragmentation}: Don’t split a single atomic task unnecessarily.
    \end{itemize}
    \item \textbf{Tool Trust Principle}: Calls to undefined tools → score 0; retries due to external failures are necessary; repeated invalid verification → redundant.
\end{enumerate}

\textbf{Special Judgments (Common Failure Cases)}
\begin{enumerate}
    \item \textbf{Parameter Responsibility}: Hallucinated or malformed parameters → score 0.
    \item \textbf{Repeated Reads}: Re-fetching already obtained info → score 0 unless justified (e.g., latest snapshot).
    \item \textbf{Invalid Retries}: Retries allowed only for transient errors (timeout, network); retries for clear errors → score 0.
    \item \textbf{Stochastic Interfaces}: Multiple calls allowed if randomness is required, with reasoning.
    \item \textbf{Task Consistency}: Substituting or simplifying the task that deviates from the goal → score 0.
\end{enumerate}

\medskip
\textbf{Evaluation Procedure}
\begin{enumerate}
    \item Identify user goal, tools, and constraints.
    \item \textbf{Parameter Source Check}: Verify origin of all parameters, detect hallucination.
    \item Step-by-step reconstruction: record context, existing info, and call intent.
    \item Per-call scoring: evaluate each call according to criteria, explain reasoning.
    \item Global summary: In the `thought` field, summarize strategy, key decisions, issues, and improvements.
\end{enumerate}

\medskip
\textbf{Output Format}
Strictly output **valid JSON only**. Example structure:

\begin{Verbatim}[breaklines=true]
{
"tool_call_num": <total number of tool calls>,
"tool_evaluations": [
    {
    "tool_index": 1,
    "tool_name": "<tool name>",
    "reasoning": "<concise justification: analyze necessity, parameter correctness, and conciseness in context>",
    "score": 0.0 | 1.0
    }
],
"thought": "<overall analysis>",
"tool_score_list": [score_1, score_2, ...]
}
\end{Verbatim}

\medskip
\textbf{Additional Notes}
\begin{itemize}
    \item Tool definitions must be strictly followed; do not infer unavailable capabilities.
    \item Parameter source checking is mandatory before evaluating any call.
    \item Maintain consistent judgment standards across similar scenarios.
    \item Strictly adhere to JSON formatting; escape quotes inside string values.
    \item Respond in Chinese.
\end{itemize}

\medskip
\textbf{Complete trajectory:}

\begin{verbatim}
{trajectory}
\end{verbatim}

\end{tcolorbox}


\begin{tcolorbox}[
    title={Example Prompt 3: Semantic Topology Extraction for Q–A Instance Synthesis},
    breakable,
    enhanced,
    colback=white,
    colframe=gray!80!black,
    before skip=0pt,
    width=\dimexpr\linewidth+2cm\relax, 
    enlarge left by=-0.75cm,
    enlarge right by=-0.75cm,
    left=2mm, right=2mm, top=2mm, bottom=2mm,
    before upper=\linespread{0.85}\selectfont
]
\textbf{Role}
You are an expert Data Architect specializing in building \textbf{Tool-Use} evaluation datasets for Large Language Models (LLMs). Your task is to generate high-quality Question-Answer (QA) pairs and their detailed reasoning decomposition paths based on the given scenarios and constraints.

\medskip
\textbf{Input Data}
Please generate data based on the following input variables:

\begin{enumerate}
    \item \textbf{Domain}: \{\{Domain\}\}
    \item \textbf{Knowledge Corpus}: \{\{Knowledge\_Corpus\}\}
    \item \textbf{Min\_Num\_Hops}: \{\{min\_num\_hops\}\}
    \item \textbf{Max\_Num\_Hops}: \{\{max\_num\_hops\}\}
    \item \textbf{Count}: \{\{num\_samples\}\}
\end{enumerate}

\medskip
\textbf{Scenario Definitions}

You need to construct \{\{num\_samples\}\} complex \textbf{User Queries} and decompose them into sub-questions. Based on the input constraints, the data must adhere to one of the following logical structures:

\begin{enumerate}
    \item \textbf{Single-Hop}: Contains exactly one sub-question.
    \item \textbf{Parallel Single-Hop}: The user query contains multiple independent sub-tasks that are mutually exclusive and can be executed in parallel.
    \item \textbf{Multi-Hop}: Contains serialized dependency relationships (sub-question $q_{i+1}$ depends on the answer $a_i$).
    \item \textbf{Parallel Multi-Hop}: A hybrid structure. It contains both independent parts that can be parallelized and parts that depend on previous results.
\end{enumerate}

\medskip
\textbf{Constraints \& Guidelines}

\begin{enumerate}
    \item \textbf{Quantity \& Diversity}: Strictly generate \{\{num\_samples\}\} data entries. Each entry must involve different entities, attributes, or specific scenarios. Avoid repetitive templates.
    \item \textbf{Realism \& Specificity}: Generated sub-questions and sub-answers must be specific (e.g., specific dates, amounts, real entity names) and not vague.
    \item \textbf{Task Parallelism}: Multi-step tasks must automatically identify and split mutually independent sub-questions (\texttt{is\_parallel = true}).
    \item \textbf{Tool-Oriented}: Each \texttt{sub\_question} must be solvable by a single atomic API tool. \texttt{sub\_answer} contains only the final result.
    \item \textbf{Clear Dependencies}: Explicitly indicate in JSON \texttt{dependency} field which step's output the current step depends on (\texttt{null} for parallel steps). \texttt{hop\_level} indicates layers.
    \item \textbf{Corpus Compliance}: Answers must be derived from the corpus; otherwise, synthesize specific reasonable data.
    \item \textbf{Hop Range}: \texttt{hop\_level} must be between \{\{min\_num\_hops\}\} and \{\{max\_num\_hops\}\}. Decide \texttt{Num\_Hops} based on \texttt{Domain} and \texttt{Knowledge\_Corpus}.
\end{enumerate}

\medskip
\textbf{Output Format}

Output \textbf{only} a JSON object list containing \{\{num\_samples\}\} objects. The format is:

\begin{Verbatim}[breaklines=true]

[
  {
    "scenario_type": "Single-Hop" | "Multi-Hop" | "Parallel Single-Hop" | "Parallel Multi-Hop",
    "main_question": "...",
    "final_answer": "...",
    "decomposition_trace": [
      {
        "_uuid": 1,
        "hop_level": 1,
        "sub_question": "...",
        "is_parallel": true,
        "dependency": null, 
        "sub_answer": "..."
      },
      ...
    ]
  },
  ...
]

\end{Verbatim}

\end{tcolorbox}

\begin{tcolorbox}[
    title={Example Prompt 4: Generating Function Code for the Synthesis Tool},
    breakable,
    enhanced,
    colback=white,
    colframe=gray!80!black,
    before skip=0pt,
    width=\dimexpr\linewidth+2cm\relax, 
    enlarge left by=-0.75cm,
    enlarge right by=-0.75cm,
    left=2mm, right=2mm, top=2mm, bottom=2mm,
    before upper=\linespread{0.95}\selectfont
]
Implement a function in accordance with a provided tool document, a set of question–answer pairs, and a given call statement. The implementation must strictly follow the tool specification and include robust, defensive error handling.

\medskip
\textbf{Procedure}
\begin{enumerate}
    \item \textbf{Review the Tool Document} \\
    Carefully parse the tool document to extract the function name and the exact parameter specification (names, types, required/default values). These must be used \emph{verbatim} in the implementation.

    \item \textbf{Interpret the Question--Answer Pairs} \\
    Use the pairs to infer:
    \begin{itemize}
        \item how problem statements map to function inputs,
        \item how outputs should be computed or formatted to match the expected answers.
    \end{itemize}

    \item \textbf{Develop the Function Implementation} \\
    Implement the function such that:
    \begin{itemize}
        \item the function call described by the call statement produces the correct result,
        \item the function name matches the tool document exactly,
        \item parameters are defined exactly as specified (including ordering and defaults where applicable),
        \item internal logic derives outputs consistent with the question--answer pairs,
        \item if any parameters have default values, ensure the function’s return value fully contains the corresponding expected answer (i.e., the answer must appear as a substring within the returned value),
        \item the function can yield diverse outputs when required (simulate additional fields or return values if implied by the tool document).
    \end{itemize}

    \item \textbf{Add Comprehensive Error Handling} \\
    Implement a reliable validation and error-reporting strategy to handle:
    \begin{itemize}
        \item incorrect parameter types,
        \item missing required parameters,
        \item invalid values or out-of-range inputs,
        \item any other foreseeable runtime issues.
    \end{itemize}
    Errors should result in clear, stable messages that help diagnose the issue without breaking execution.
\end{enumerate}

\medskip
\textbf{Output Requirements}

Return \textbf{only} a JSON object (no other text) using the following structure:
\begin{itemize}
    \item \texttt{"analysis"}: A detailed explanation of the implementation approach, including design rationale for parameters, output formatting, and exception/error-handling logic.
    \item \texttt{"function"}: The full function implementation, including code and explanatory comments.
\end{itemize}

\medskip
\textbf{Notes}

\begin{itemize}
    \item Parameter names and types must match the tool document \emph{exactly}.
    \item Only Python~3 built-in libraries may be used.
    \item If the tool specification implies additional return values, simulate them in a way consistent with the documentation.
    \item Error handling should be exhaustive and anticipatory.
    \item Ensure that for any input \( q \), the produced result is uniquely \( a \).
\end{itemize}

\medskip
\textbf{Provided Inputs}

\textbf{Tool Document}
\begin{verbatim}
{document}
\end{verbatim}

\textbf{Question--Answer Pairs}
\begin{verbatim}
{pairs}
\end{verbatim}

\textbf{Call Statement}
\begin{verbatim}
{call_statement}
\end{verbatim}

\end{tcolorbox}



\end{document}